\documentclass{article}

\usepackage{microtype}
\usepackage{graphicx}
\usepackage{subfigure}
\usepackage{booktabs} 
\usepackage{xspace}

\usepackage{amsmath,amssymb}
\usepackage{threeparttable}
\usepackage{multicol,multirow}
\usepackage{balance}

\newcommand{\ourmethod}{AdaXpert\xspace}

\def\ie{i.e.,~}
\def\eg{e.g.,~}

\def\blue{\textcolor{black}}

\def\mata{\textcolor{black}}

\def\xu{\textcolor{black}}
  \def\mA{{\mathcal A}}
  
  \def\mC{{\mathcal C}}
  \def\mD{{\mathcal D}}

  \def\mM{{\mathcal M}}
  \def\mN{{\mathcal N}}
  
  \def\mP{{\mathcal P}}
  
  \def\mR{{\mathcal R}}
  \def\mS{{\mathcal S}}

  \def\mV{{\mathcal V}}
  \def\mW{{\mathcal W}}
  \def\mX{{\mathcal X}}

  \DeclareMathAlphabet\mathbfcal{OMS}{cmsy}{b}{n}

  \def\0{{\bf 0}}
  \def\1{{\bf 1}}

  \def\bM{{\bf M}}

  \def\mmR{{\mathbb R}}

  \def\blue{\textcolor{black}}

  \def\citep{\cite}

  \def \mbP {\mathbb{P}}

\def\blue{\textcolor{black}}
\def\rh{\textcolor{black}}
\def\grammar{\textcolor{black}}

\usepackage{hyperref}

\usepackage[accepted]{icml2021}

\icmltitlerunning{AdaXpert: Adapting Neural Architecture for Growing Data}

\begin{document}

\twocolumn[
\icmltitle{AdaXpert: Adapting Neural Architecture for Growing Data}




\icmlsetsymbol{equal}{*}


\begin{icmlauthorlist}
\icmlauthor{Shuaicheng Niu}{equal,scut,edu}
\icmlauthor{Jiaxiang Wu}{equal,tencent}
\icmlauthor{Guanghui Xu}{scut}
\icmlauthor{Yifan Zhang}{nus}\\ 
\icmlauthor{Yong Guo}{scut}
\icmlauthor{Peilin Zhao}{tencent}
\icmlauthor{Peng Wang}{npu}
\icmlauthor{Mingkui Tan}{scut,pz}
\end{icmlauthorlist}

\icmlaffiliation{scut}{School of Software Engineering, South China University of Technology, China}
\icmlaffiliation{pz}{Pazhou Laboratory, China}
\icmlaffiliation{tencent}{Tencent AI Lab, China}
\icmlaffiliation{nus}{National University of Singapore, Singapore}
\icmlaffiliation{npu}{Northwestern Polytechnical University, China}
\icmlaffiliation{edu}{Key Laboratory of Big Data and Intelligent Robot, Ministry of Education, China}

\icmlcorrespondingauthor{Mingkui Tan}{mingkuitan@scut.edu.cn}

\icmlkeywords{Machine Learning, ICML}
\vskip 0.3in ]



\printAffiliationsAndNotice{\icmlEqualContribution} 


\begin{abstract}
\blue{In real-world applications, data often come in a growing manner, where the data volume and the number of classes may increase dynamically. This will bring a critical challenge for learning: given the increasing data volume or the number of classes, one has to instantaneously adjust the neural model capacity to obtain promising performance. Existing methods either ignore the growing nature of data or seek to independently search an optimal architecture for a given dataset, and thus are incapable of promptly adjusting the architectures for the changed data.  To address this, we present a neural architecture adaptation method, namely \textbf{Ada}ptation e\textbf{Xpert} (\ourmethod), to efficiently adjust previous architectures on the growing data. Specifically, we introduce an architecture adjuster to generate a suitable architecture for each data snapshot, based on the previous architecture and the different extent between current and previous data distributions. Furthermore, we propose an adaptation condition to determine the necessity of adjustment, thereby avoiding unnecessary and time-consuming adjustments. Extensive experiments on two growth scenarios (increasing data volume and number of classes) demonstrate the effectiveness of the proposed method.}
\end{abstract}

\section{Introduction}

\begin{figure}[t]
\centering
\begin{minipage}[t]{0.24\textwidth}
\centering
    \includegraphics[height=0.7\textwidth]{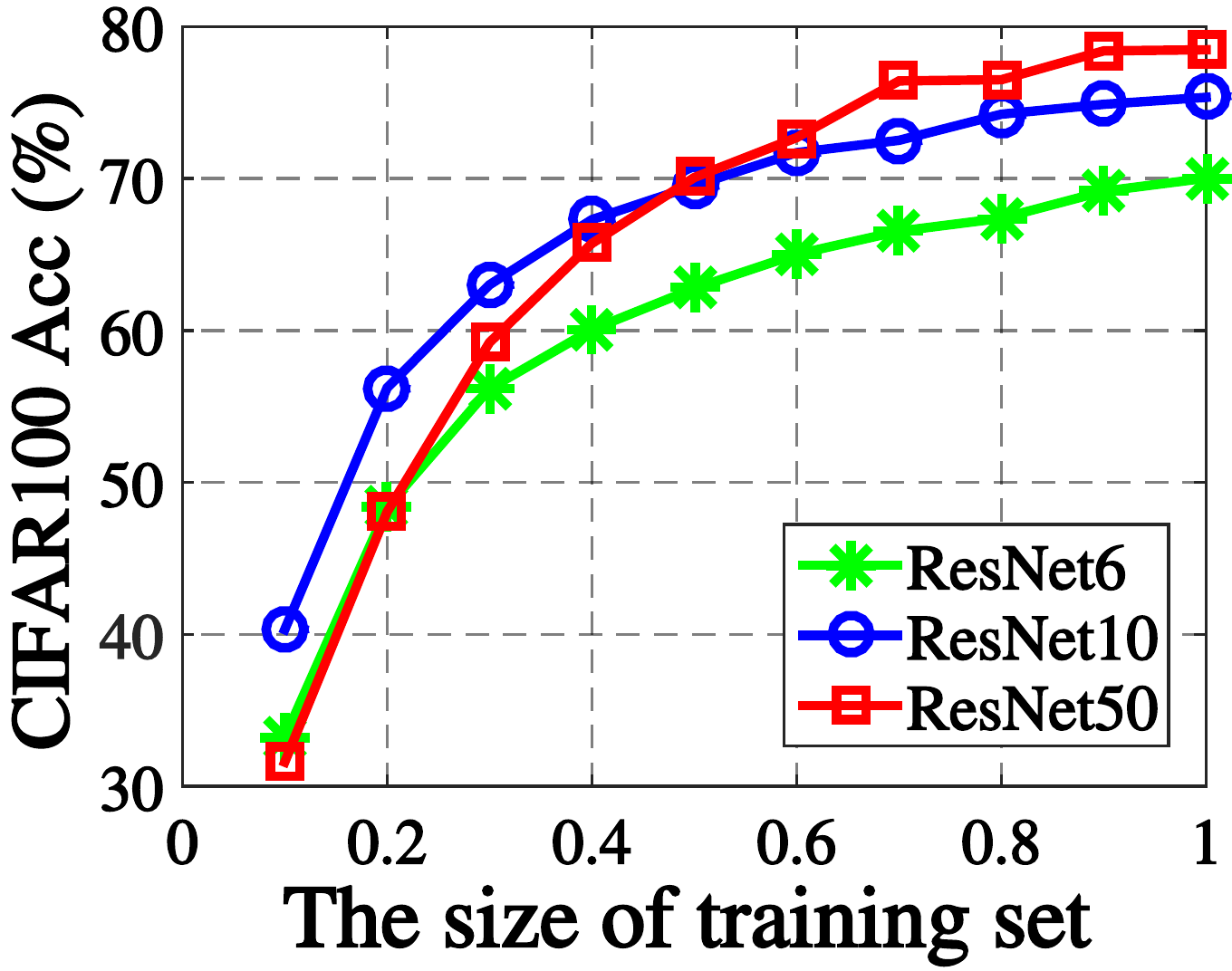}
\end{minipage}
\hspace{-0.1in}
\begin{minipage}[t]{0.24\textwidth}
\centering
    \includegraphics[height=0.7\textwidth]{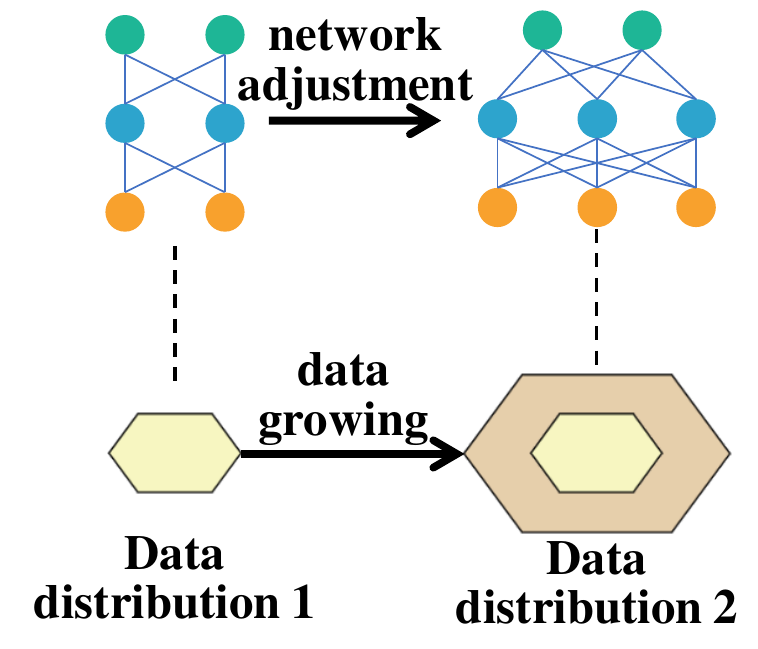}
\end{minipage}
\vspace{-0.05in}
\caption{The motivation for architecture adaptation. \textbf{Left}: Performance comparisons of ResNets trained on different subsets of CIFAR100. The optimal architecture varies among different subset sizes.
\textbf{Right}: Since data come in a growing manner and the data distribution may dynamically change, one should adjust the model architecture according to the shift of data distribution.}
\vspace{-0.1in}
\label{fig:fitting_ability}
\end{figure}

\mata{Deep neural networks (DNNs) have achieved state-of-the-art results in many challenging tasks, including image classification~\cite{hu2018squeeze,Lu2021NeuralAT}, neural language processing~\cite{devlin2019bert,brown2020language}, and many other areas~\cite{cao2019multi,zhang2020collaborative,guo2020closed,zeng2020dense}.} One of the key factors behind the success of DNNs lies in the design of effective neural architectures, including 1) the manually designed architectures such as ResNet~\cite{resnet} and MobileNet~\cite{howard2017mobilenets}; \grammar{2) the} automatically designed architectures such as~\cite{zoph2018learning,cai2018proxylessnas,tan2019mnasnet}. However, these methods often design a fixed architecture for a specific task/dataset. 

In real-world applications, data often come in \blue{a growing} manner. For example, 
intelligent edge devices (\eg billions of mobile phones and surveillance cameras) and medical imaging devices continue collecting new data every day~\cite{Grantz2020TheUO,Liang2019EvaluationAA}. Specifically, the \grammar{newly} collected data have the following two types: (1) \blue{increasing data volume: the labels of new data have appeared in previous data, and the growing do not change the label space of data; (2) increasing number of classes: the \grammar{newly} arriving data have different labels from previous data, and thus the label space of data is growing.} For both two scenarios, the data distribution may dynamically change. 
Since the optimal network architecture may vary under different data distributions~\cite{zoph2016neural},  
when applying DNNs to growing data, one can (and should) dynamically adjust the architecture for better performance (see Figure~\ref{fig:fitting_ability}).

To achieve the above goal, one straightforward solution is to redesign a network architecture when new data arrive. However,
the design of effective neural architectures substantially relies on human expertise. Moreover, human design cannot fully explore the complete architecture space, resulting in sub-optimal architectures~\cite{zoph2016neural}. Beyond manual design, one can also resort to automatic neural architecture search (NAS) techniques~\cite{cai2018proxylessnas,tan2019mnasnet}.
Nevertheless, such methods design a new architecture for each instance of data growth separately and from scratch, and ignore that previous \blue{architectures} are transferable, leading to inferior design efficiency. 
Moreover, neither manual nor automatic design considers the necessity of architecture adjustment to further improve the adaptation efficiency. Intuitively, it is unnecessary to conduct adaptation if newly arrived data are very similar to previous data.

To address the above limitations, we propose a neural architecture adaptation method, called \ourmethod (\textbf{Ada}ptation e\textbf{Xpert}), which consists of an architecture adjuster and an adaptation condition. 
Specifically, we first adopt the Wasserstein distance to quantitatively measure the difference between current and previous data. Then, 
the adjuster takes the previous architecture and data difference as inputs, and generates a suitable architecture for the current data. Next, the adjuster receives a reward for this adjustment, where the reward is designed to compromise between the accuracy and computational efficiency.
The final adjusted architecture is generated by a well-trained adjuster. 
It is worth mentioning that our method aims to find a new optimal architecture for current data rather than simply expand the previous architecture to a larger one. The architecture adjustment is elastic, \eg the adjustment may remove redundant layers from the previous architecture and increase the capacity (kernel size and/or number of channels) of certain layers, as illustrated in Figure~\ref{fig:arch_diff}.
Moreover, we propose an adaptation condition to determine the necessity of architecture adjustment. In this way, we avoid unnecessary adjustments if the newly arrived data are highly similar to the previous one, thereby further improving the adjustment efficiency.
Based on the above considerations, our \ourmethod~is able to adjust an architecture automatically to obtain better performance on the current data with as minimal computational cost as possible.
Our main contributions are summarized as follows:
\begin{itemize}
    \item We propose a network adaptation method for growing data. By considering the difference between the current and previous data, our method adaptively adjusts the model architecture to achieve better performance while maintaining small computational cost.
    \item We propose an adaptation condition to determine the necessity of architecture adjustment. With this condition, our method avoids unnecessary adaptation for highly similar data.
    \item Experiments on two data growth scenarios, \ie increasing data volume and the number of classes, demonstrate the effectiveness and superiority of our method.
\end{itemize}

\section{Related Work}

\textbf{Neural architecture search} (NAS)
has attracted increasing attention to automatically design effective architectures.
The classical NAS problem~\cite{zoph2016neural} exploits the paradigms of reinforcement learning (RL) to generate the model descriptions of DNNs.
RL-based methods~\cite{pham2018efficient,tan2019mnasnet,zoph2018learning} seek to learn a controller with a policy to generate architectures. Beyond RL, evolutionary-based~\cite{real2018regularized,DBLP:journals/corr/abs-1910-06961} and gradient-based~\cite{liu2018darts,xu2020pcdarts} algorithms also discover new architectures with excellent performance. 
Recently, meta learning-based methods~\cite{Lian2020TowardsFA,Wang2020MNASMN} focus on the few-shot problem and automatically learn a meta-architecture that is intended to adapt to new tasks quickly. 
\mata{Unlike NAS that design a fixed architecture, we dynamically adjust the network architecture to handle the problem of growing data.}

\textbf{Continual learning} (CL)
aims to transfer the knowledge learned from previous tasks to future scenarios. 
To solve a new task, replay-based methods~\cite{Rebuffi2017iCaRLIC,Chaudhry2019ContinualLW,Rolnick2019ExperienceRF} selectively store samples of previous tasks for the training on new tasks.
To mitigate the catastrophic forgetting issue, regularization-based methods~\cite{Kirkpatrick2017OvercomingCF,Liu2018RotateYN,Lange2020UnsupervisedMP} introduce a regularization term in the loss function, which requires the model to not change important parameters of previous tasks. 
Another parallel task is online learning (OL)~\cite{Hoi2018OnlineLA,zhang2019online}, which aims to learn a well-performed model based on a sequence of training samples. 
\mata{With the data growth, CL aims to overcome the forgetting issue on previous data and conduct adjustments to achieve better performance on new data, and OL focuses on the parameter learning of a certain model. In this work, we seek to instantaneously adjust the architecture on the entire dataset for each time of data growth.}


\textbf{Progressive neural networks}.
To improve the model capacity, CL methods~\cite{Rusu2016ProgressiveNN,Xu2018ReinforcedCL,Rosenfeld2020IncrementalLT} propose to dynamically expand their network architectures. 
These methods fix the layers of previous tasks and grow branches for new tasks. 
Moreover, DEN~\cite{Yoon2018LifelongLW} first expands the architecture to a large size for a new task, and then use a pruning method to remove the unimportant weights. Recently, \cite{Gao2020EfficientAS} and ~\cite{Li2019LearnTG}
combine NAS techniques to design architectures for each task to achieve the goal of CL.
However, these methods ignore the distribution difference between the current and previous data, and thus are hard to determine a suitable model size of the adjusted architecture. 
\mata{In this work, we dynamically adjust the architecture based on previous architectures and the properties of growing data. Moreover, beyond the expansion, our adjustment may remove redundant layers or add new layers.}


\begin{figure*}[t]
    \centering
    \includegraphics[width=1.0\textwidth]{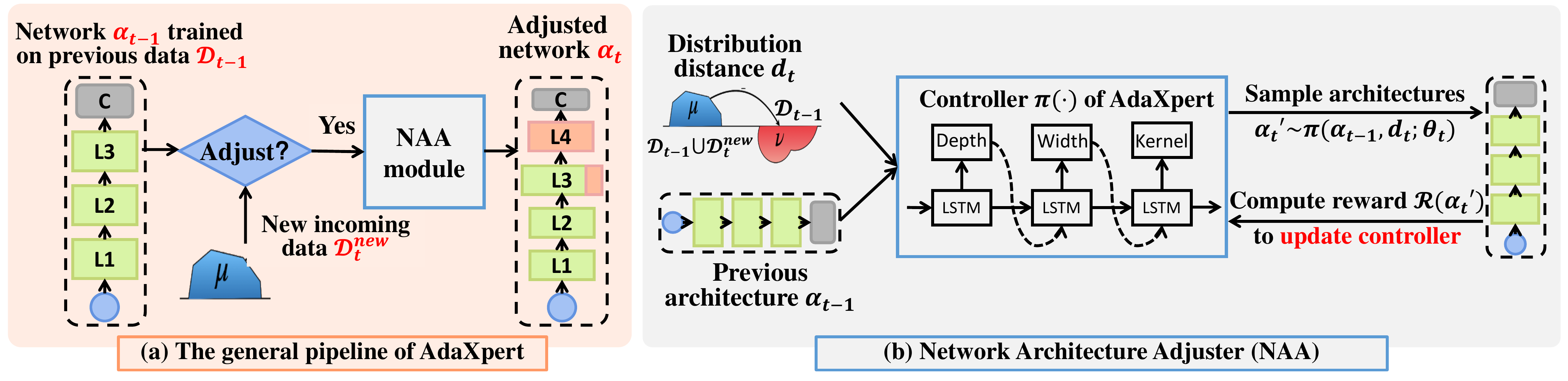}
    \vspace{-0.15in}
    \caption{An illustration of our proposed \ourmethod. (a) At time-step $t$, given new incoming data $\mD_t^{new}$ and a previous model $\alpha_{t-1}$, we first determine whether to adjust the architecture. If necessary, the $\alpha_{t-1}$ is fed into the NAA module for architecture adaptation.
    (b) Our controller takes the architecture of $\alpha_{t-1}$ and the distribution distance between current data $\mD_{t-1}\cup\mD_t^{new}$ and previous data $\mD_{t-1}$ as inputs, and outputs an adjusted architecture. The controller will then obtain a reward and thus can be trained via policy gradient methods. Last, we adopt the well-trained NAA to generate the final adjusted architecture $\alpha_t$.
    }
    \label{fig:overall}
\end{figure*}

\section{Proposed Method}

\subsection{Problem Definition}
In this paper, we aim to dynamically adjust neural network architectures along with the growth of data.
Formally, we denote a series of new incoming data as  $\mD_t^{new}\small{=}\{(x_i^t,y_i^t)\}_{i=1}^{n_t},$ $t\small{=}1,...,T$, where $x_i^t\small{\in}\mX, y_i^t\small{\in}\{1,...,C_t\}$, $\mX$ is the input image space, $n_t$ and $C_t$ are the number of images and classes, respectively. Moreover, we denote all the accumulated data at time-step $t$ as $\mD_t=\mD_1^{new}\cup\cdots\cup\mD_t^{new}$.

For the above growing datasets $\{\mD_t\}_{t=1}^{T}$, their corresponding data distributions may dynamically change, and thus the optimal architecture for different $\mD_t$ may also vary. However, existing methods usually design a fixed architecture for a specific task/dataset, while a single fixed network architecture may not be optimal for all $\{\mD_t\}_{t=1}^{T}$. To achieve better performance, one should design different architectures for different $\mD_t$, \ie dynamically adjust the model architecture along with the growth of data.  

To achieve the above goal, an intuitive way is to perform  neural architecture search for each $\mD_t$ separately, to obtain a corresponding architecture $\alpha_t$. However, this method ignores that the previous $\alpha_{t-1}$ are transferable for $\mD_t$, which can improve the search efficiency of $\alpha_t$. 
Moreover, it is also important to avoid unnecessary search processes when the new incoming data are very similar to the previous data.

To address the above challenges, we propose a neural architecture adaptation method, namely \textbf{Ada}ptation e\textbf{Xpert} (\ourmethod), which 
aims to automatically design dynamic networks with minimal expert intervention. The \ourmethod mainly consists of two key components: (1) \blue{a} reinforcement learning based neural architecture adjuster (see Sect.~\ref{sec:arch_adapt}), which adaptively adjusts the previous model architecture $\alpha_{t-1}$ to a new $\alpha_t$ according to the properties of new incoming data $\mD_{t}^{new}$; (2) \blue{an} adaptation condition that determines the necessity of architecture adjustment (see Sect.~\ref{sec:discrimination}), thereby avoiding unnecessary adjustments. The detailed pipeline of \ourmethod is summarized in Algorithm~\ref{alg:overall}.

\begin{figure*}[t]
	\begin{minipage}[t]{0.48\linewidth}
		\centering
		\begin{algorithm}[H]
        \small
	\caption{The overall algorithm of \ourmethod.}
	\label{alg:overall}

	\begin{algorithmic}[1]
    \REQUIRE{Incoming datasets $\{\mD_t^{new}\}_{t=1}^{T}$; well-trained model $\alpha_1$ for $\mD_1^{new}$; supernet $\mN_1$ and controller $\pi(\cdot;\theta_1)$;
    threshold $\epsilon$.}
	\STATE Let $\mD_1=\mD_1^{new}$.
    \FOR{t=2,...,T}
        \STATE Let $\mD_t\small{=}\mD_{t-1}\small{\cup}\mD_t^{new}$, $\mN_t\small{=}\mN_{t-1}$ and $\theta_t\small{=}\theta_{t-1}$. 
        \STATE Compute accuracy difference $H_t$ using Eqn.~(\ref{eq:discrimination}).
    \IF{$H_t > \epsilon$}{
    \STATE Update $\mN_t$ and $\pi(\cdot;\theta_t)$ on $\mD_t$ using Algorithm~\ref{alg:training}.
    \STATE Generate the adjusted architecture $\alpha_t \sim \pi(\cdot; \theta_t)$.
    \STATE Retrain the adjusted architecture $\alpha_{t}$ on $\mD_t$.
    }
    \ELSE 
    \STATE Let $\alpha_t=\alpha_{t-1}$.
    \ENDIF
    \ENDFOR
    \ENSURE The adjusted architectures $\{\alpha_t\}_{t=1}^{T}$.
 
	\end{algorithmic}
	\end{algorithm}
		\end{minipage}\hfill
	\begin{minipage}[t]{0.48\linewidth}
			\centering
			\begin{algorithm}[H]
            \small
	\caption{Training of Network Architecture Adjuster.}
	\label{alg:training}
	\begin{algorithmic}[1]
	    \REQUIRE{Datasets \{$\mD_{t-1}, \mD_{t}\}$; previous architecture $\alpha_{t-1}$, supernet $\mN_t$ and controller $\pi(\cdot)$ with the parameters $\theta_t$;
	    hyper-parameters $\eta$  and $M$.}
	    \vspace{2pt}
	   \STATE Split $\mD_t$ into training and validation sets \{$\mD_{train}$, $\mD_{val}$\}.
	   \vspace{2pt}
        \STATE Fine-tune $\mN_t$ on $\mD_{train}$. 
        \vspace{2pt}
        \STATE Compute the WD ($d_t$) between $\mD_{t-1}$ and $\mD_t$ using Eqn.~(\ref{eq:w_distance});
        \STATE //~\emph{train the controller model}
        \vspace{1pt}
        \FOR{i=1,...,M}
        \STATE Sample  $\alpha_t'\small{\sim}\pi(\alpha_{t-1},d_t;\theta_{t})$.
        \vspace{1pt}
        \STATE Sample a batch of data from $\mD_{val}$.
        \vspace{1pt}
        \STATE Compute reward $\mR(\alpha_t')$ based on $\mN_t$ using Eqn.~(\ref{eq:reward}).
        \STATE Update $\theta_t\leftarrow \theta_t + \eta\mR(\alpha_t')\nabla_{\theta_t}\text{log}\pi(\cdot)$.
        \vspace{1pt}
        \ENDFOR
        \ENSURE  Supernet $\mN_t$ and controller $\pi(\cdot; \theta_t)$.
	\end{algorithmic}
    \end{algorithm}
	\end{minipage}
	\end{figure*}

\subsection{\rh{Dynamic Neural} Architecture Adaptation}\label{sec:arch_adapt}
Given a previous deep model and new incoming data, we aim to automatically adjust the model architecture to achieve better performance while retaining small model computational cost (\eg MAdds).
To this end, we devise a Network Architecture Adjuster (NAA) algorithm, which aims to conduct different adjustment strategies based on 
the distribution difference 
between $\mD_{t-1}$ and $\mD_t$. Specifically, if new data are very similar to the previous one, we only need to conduct a slight adjustment, \ie keeping the MAdds growth small for the adjusted architecture. Otherwise, we allow a relatively large MAdds growth for the adjustment.

In the following, we address two key problems for architecture adaptation: 1) how to measure the difference between the current data $\mD_t$ and the previous data $\mD_{t-1}$; and 2) how to design the architecture adjuster.

\textbf{Quantitative measurement of data difference.}
Different architecture adaptation strategies should be conducted based on the different extent between the current data and previous data.
To quantify such difference, we compute the distribution distance between current and previous data as follows. 

Formally, given the current dataset $\mD_t$ and the previous dataset $\mD_{t-1}$, we first feed these two datasets to the previous model $\alpha_{t-1}$ to obtain their feature embeddings $\bM_{t}\in\mmR^{m\times q}$ and $\bM_{t-1}\in\mmR^{n\times q}$, respectively. Here, $m$ and $n$ denote the number of samples in $\mD_{t}$ and $\mD_{t-1}$ respectively, and $q$ denotes the feature dimension. Then, $\bM_t$ and $\bM_{t-1}$ can be considered as two sample matrices that are sampled from two unknown distributions $\mbP_{t}$ and $\mbP_{t-1}$. To compute the distance between $\mbP_{t}$ and $\mbP_{t-1}$, one can use non-parametric estimation methods to compute the Kullback–Leibler (KL) divergence~\cite{nguyen2007nonparametric} or Wasserstein distance (WD)~\cite{sriperumbudur2010non}. 

However, the above non-parametric estimation methods may be computationally expensive. For example, computing the KL divergence needs solving a quadratic programming problem. Luckily, our preliminary studies show that the sample matrices $\bM_t$ and $\bM_{t-1}$ approximately satisfy the multivariate Gaussian distribution (more details are put in the supplementary). Therefore, in this paper, we assume that $\mbP_t$ and $\mbP_{t-1}$ are two multivariate Gaussian distributions, and use the Maximum Likelihood Estimation method to obtain their distribution parameters, \ie $\mbP_t\sim\mN(\mu_t,\Sigma_t)$ and $\mbP_{t-1}\sim\mN(\mu_{t-1},\Sigma_{t-1})$. Then, we compute the Wasserstein distance~\cite{takatsu2011wasserstein} as follows:
\begin{align}
\label{eq:w_distance}
     \mW(\mD_t,\mD_{t-1}) =& ||\mu_t-\mu_{t-1}||_2^2+ \nonumber \\  &\text{tr}\Big(\Sigma_t\small{+}\Sigma_{t-1}\small{-}2(\Sigma_{t-1}^{1/2}\Sigma_t\Sigma_{t-1}^{1/2})^{1/2}\Big).
\end{align}
Here, one can also use other metrics to compute the distribution distance, such as KL and Jensen-Shannon divergence~\cite{Fuglede2004Jensen}.
\mata{More discussions about WD are put in supplementary.}
Based on $\mW(\mD_t,\mD_{t-1})$, we devise a data difference-aware controller to conduct different architecture adjustments. 
Due to the highly non-convex nature of our adjustment problem, we cast it into a Markov Decision Process (MDP),  and then train the controller using reinforcement learning methods.

\textbf{MDP reformulation for Neural Architecture Adjuster~(NAA).}
Since the architecture adjustment process is essentially a multi-step decision making process, we formalize the adjustment process as \grammar{an} MDP. Formally, \grammar{the} MDP can be defined as a tuple $\mM \small{=} (\mS, \mA, \mP, \mR)$, where $\mS$ is a finite set of states, $\mA$ is a finite set of actions, $\mP: \mS \small{\times} \mA \small{\rightarrow} \mS$ is the state transition distribution, $\mR : \mS \small{\times} \mA \small{\rightarrow} \mR$ is the reward function. Moreover, a policy $\pi_\theta$ determines an action given the current state. In the context of NAA, as illustrated in Figure~\ref{fig:overall} for time-step $t$, we denote state as $s=[\alpha_{t-1}, d_t]\in \mS$, where $d_t=\mW(\mD_t, \mD_{t-1})$ is the distribution difference between the current and previous data. Given such a state $s$, a policy (controller) takes a series of actions $a=\pi_\theta(s)\in\mA$ to determine each layer's operation of the adjusted architecture $\alpha_t'$. Formally, the action space is defined based on different types of architecture search space. Then, the controller receives a reward $r=\mR(\alpha_t')$. More details about the reward design can be found in Sec.~\ref{sec:reward}.

\textbf{Training of NAA.} The goal of our NAA is to maximize an expectation reward $\mathbb{E}[\mR(\alpha)]$, represented by
solving the following optimization problem:
\begin{align}	
    \label{eq:loss_pg}
	\max_{\theta}~ \mathbb{E}_{\pi_\theta}[\mR(\alpha)].
\end{align}
Following policy gradient methods~\cite{williams1992simple,schulman2017proximal}, 
we update $\theta$  by ascending the gradient:
\begin{align}
    \theta \leftarrow \theta + \eta\mR(\alpha)\nabla_{\theta}\text{log}\pi_\theta(\alpha).
\end{align}
The training details of NAA are summarized in Algorithm~\ref{alg:training}.

\subsection{When to Adapt Network Architecture}\label{sec:discrimination}

Given a previous model $\alpha_{t-1}$ and the current data $\mD_t$, it should be considered whether the model architecture needs adjustment.
For example, given an optimal architecture $\alpha$ that searched on the MNIST dataset, even if we receive more incoming MNIST images, it is unnecessary to adjust $\alpha$ since it is already optimal. This indicates the necessity to consider the previous architecture's feasibility when performing architecture adaptation. 
Therefore, as shown in Figure~\ref{fig:overall} (a), for each time new data arrives, we use an adaptation condition to determine the necessity of adjustment, and therefore improve the adaptation efficiency.

Formally, at time-step $t$, given the previous model $\alpha_{t-1}$, current data $\mD_t$ and previous data $\mD_{t-1}$, we compute the following accuracy difference for further decision:
\begin{equation}
    \label{eq:discrimination}
    H_t =\Phi(\mD_{t-1}; \alpha_{t-1})-\Phi(\mD_t;\alpha_{t-1}),
\end{equation}
where $\Phi(\mD;\alpha)$ is some performance metric of model $\alpha$ on dataset $\mD$. For classification models, we choose  top-1 accuracy as the metric. 
Note that for the growing data with new class labels, the previous model will make wrong predictions, since the classifier can not predict the new labels. Based on the above accuracy difference, one can determine whether to adjust the previous architecture $\alpha_{t-1}$. 
Specifically, given $H_t$ and its threshold $\epsilon$, we only adjust the model architecture when $H_t >\epsilon$. 

In our method, we exploit WD to measure the extent of data difference, and adopts accuracy difference as the adaptation condition of architecture adjustment. The reasons are as follows: (1) For adaptation condition, the accuracy difference is more intuitive for humans. In contrast, since WD is a distribution distance, it is hard to set a suitable threshold for WD to determine the adjustment necessity. (2) To recognize difference extent of data, WD has a stronger discrimination ability than the accuracy difference. Specifically, for varying label space, the accuracy difference is more determined by the number of new data, while WD is computed according to the underlying properties of the data itself. 
 
\subsection{Reward Design for NAA} 
\label{sec:reward}
The reward function $\mR(\cdot)$ is very important for training our NAA model. In this subsection, we will provide our rewards function.
 For simplicity, we only illustrate the training of NAA for a single round, and in other rounds, the NAA can be trained in the same way. 
Given the current and previous datasets $\mD_t$ and $\mD_{t-1}$, we denote the WD between them as $d_t=\mW(\mD_t,\mD_{t-1})$. 
Then, for the current state $s$, the NAA takes a series of actions to obtain an adjusted architecture $\alpha_t'$. The final reward is computed as follows:
\begin{align}\label{eq:reward}
    \mR(\alpha_t')\small{=}&\mV(\alpha_t')\small{-}\mV(\alpha_{t\small{-}1})  
              \small{-}  \frac{\lambda}{d_t}\big(\mC(\alpha_t')\small{-}\mC(\alpha_{t\small{-}1})\big),
\end{align}
where $\mV(\alpha)$ and $\mC(\alpha)$ denote the validation accuracy and computational complexity of model $\alpha$ respectively, and $\lambda$ is a trade-off parameter. We adopt MAdds as our metric to measure the computational complexity of $\alpha$, and one can also use other metrics, \eg the inference latency.

To obtain the validation accuracy  $\mV(\alpha)$, one can train it from scratch and then validate it on the validation set. However, this will result in unbearable computational burdens. In this paper, we exploit a weight sharing technique~\cite{pham2018efficient} to construct a super network, \ie a large computational graph, where each network architecture shares 
parameters. In this sense, once the super network is trained, all architectures inherit their weights directly from the super network, and then use these weights for further evaluation.

For the first item $\mV(\alpha_t')\small{-}\mV(\alpha_{t-1})$ in Eqn.~(\ref{eq:reward}), we hope that the adjusted architecture $\alpha_t'$ would achieve higher validation accuracy than the original $\alpha_{t-1}$. 
For the second item $\frac{\lambda}{d_t}\times(\mC(\alpha_t')-\mC(\alpha_{t-1}))$ in Eqn.~(\ref{eq:reward}), architectures with excessive computational cost are penalized,
\mata{and the $d_t$ is used to adaptively regularize the magnitude of the adjusted architecture. If $d_t$ is small, \ie the incoming data are very similar to the original, the reward function will devote more attention to constraining the computational complexity of the adjusted architecture.}

\begin{table*}[t]
	\caption{\textbf{Scenario~I}: Comparison on \xu{ImageNet-100} with different sizes of training set. We report Acc. (\%, $\uparrow$) and \#MAdds (Million, $\downarrow$).}
	\label{tab:scenario_1}
\newcommand{\tabincell}[2]{\begin{tabular}{@{}#1@{}}#2\end{tabular}}
 \begin{center}
 \begin{threeparttable}
    \resizebox{0.78\textwidth}{!}{
 	\begin{tabular}{lcc|cc|cc|cc|cc}\toprule
        \multirow{2}{*}{\tabincell{c}{Methods}}
        &\multicolumn{2}{c|}{\mata{10\% training set}}&\multicolumn{2}{c|}{20\% training set}&\multicolumn{2}{c|}{40\% training set}&\multicolumn{2}{c|}{80\% training set}&\multicolumn{2}{c}{100\% training set}\cr\cmidrule{2-3}\cmidrule{4-5}\cmidrule{6-7}\cmidrule{8-9}\cmidrule{10-11}
        & Acc. &MAdds & Acc. &MAdds &Acc. &MAdds &Acc. &MAdds &Acc. &MAdds \cr
        \midrule
         MobileNetV2 &52.72	&300 & 64.02	 	& 300 & 72.08	& 300  &78.36	 	& 300 	& 79.60	& 300   \\
         MobileNetV2 (1.4$\times$) &54.36 &560 &64.82	 	& 560  	& 73.02	&   560     &78.88	 	& 560 & 80.76	& 560  \\
         ResNet18 &52.72 &1,814& 62.74 &	1,814  & 71.54	& 1,814   &77.54	& 1,814   & 79.30	& 1,814    \\	
         ResNet50 &38.86 &4,087 &53.26 & 4,087   & 66.76	& 4,087   &78.62	 	& 4,087  	& 80.30	& 4,087    \\
         \midrule
         MnasNet-A1 &50.30 &323 &61.14 & 323  & 71.46	& 323  & 78.82 & 323  & 79.66 & 323  \\
         EfficientNet-B0 &50.56	&398 &62.90 & 398  & 72.32	& 398  & \textbf{79.42} & 398  & 80.38 & 398   \\
         Meta-NAS &51.00 &559 &60.00 & 559  & 69.30	& 559  & 77.08 & 559  & 77.48 & 559   \\
         \midrule
         \mata{D-EfficientNets} &49.08	&\textbf{145} &61.42 & 178  & 71.28	& 203  & 78.26 & 229  & 80.44 & 278   \\
         Progressive NN &54.60 & 149 &61.72 & 181  & 71.38	& 203  & 78.78 & 244  & 80.02 & 261   \\
         DEN &54.60 & 149 &63.50 &258  & 72.20	& 315  & 78.85 & 439  & \textbf{80.84} & 515   \\
         \midrule
         \ourmethod~(ours) &\textbf{54.60} & 149 & \textbf{64.90} & \textbf{171}  & \textbf{73.28}	& \textbf{199}  & 79.28 & \textbf{232} &  80.74 & \textbf{252}  \\
        \bottomrule
	\end{tabular}}
	 \end{threeparttable}
	 \end{center}
	 \vspace{-0.1in}
\end{table*}

\section{Experiments}

In this section, we evaluate our \ourmethod with respect to two data growth scenarios,
\ie data volume growth within the same label space (\textbf{Scenario~I}) and \blue{increasing} label space (\textbf{Scenario~II}). Afterward, we conduct ablation studies to verify the effectiveness of each component in our method. Lastly, we compare architectures obtained by our adaptation procedure with those obtained by existing methods.
Code is available at \href{https://github.com/mr-eggplant/adaxpert0}{https://github.com/mr-eggplant/adaxpert0}.

\subsection{Experimental Settings}

\textbf{Datasets:} We conduct our experiments on ImageNet, a large-scale image classification dataset \cite{deng2009imagenet}. Based on ImageNet, we simulate two data growth scenarios to verify the effectiveness of our proposed method. \xu{For convenience, we denote ImageNet-\# as a subset of ImageNet, where `\#' denotes the number of classes. For instance, ImageNet-100 contains samples of the first 100 classes of the entire ImageNet.
We also name our dynamically adjusted architectures in a similar manner, \eg \ourmethod-20 denotes the architecture obtained on ImageNet-20.}

\textbf{\rh{Search space for architecture adaptation}:}
Here, we consider the architecture space based on the inverted Mobile Block~\cite{howard2019searching}.
To be specific, 
the model is divided into 5 units with gradually reduced feature map spatial size and increased number of channels.
Each unit consists of 4 layers at most, where only the first layer has stride 2 if the feature map size decreases, and all the other layers have stride 1.
In our experiments, we search for the number of layers in each unit (chosen from $\{2,3,4\}$), the kernel size in each layer (chosen from $\{3,5,7\}$), and the width expansion ratio in each layer (chosen from $\{3,4,6\}$).

\textbf{Compared methods:} 
We compare our \ourmethod with three categories of methods. (1) Manually designed networks, including MobileNetV2, MobileNetV2 (1.4$\times$)~\cite{howard2017mobilenets}, ResNet18, and ResNet50~\cite{resnet}. During the entire data growth process, these models are trained and evaluated with the same fixed architecture. (2) 
Neural architecture search (NAS) methods. EfficientNet~\cite{tan2019efficientnet} and MnasNet~\cite{tan2019mnasnet} are searched on inverted Mobile Block search space (as ours) and achieve state-of-the-art performance.
We also compare our method with Meta-NAS~\cite{shaw2018meta}, which first searches a meta architecture  on multiple tasks and then adapts it to ImageNet. 
(3) Dynamic neural networks. Progressive NN~\cite{Rusu2016ProgressiveNN} and DEN~\cite{Yoon2018LifelongLW} are two methods that dynamically adjust the network architecture from small to large with the growth of data. 
\mata{D-EfficientNets is a width-multiplier method, where the networks are re-scaled with different widths of EfficientNet-B0~\cite{tan2019efficientnet} to adapt the corresponding data.}
Please refer to the supplementary for further implementation details.

\subsection{I: Growing Data \blue{with} Same Label Space}
In this section, we conduct experiments on the data growth scenario in which the data volume is growing while the label space remains the same. 

\textbf{Simulation of growing data:} 
We simulate the data volume growth scenario on ImageNet-100 due to the high computational cost of evaluating all considered architectures on ImageNet-1000. To be specific, the data come with different ratios, \ie \{10\%, 20\%, 40\%, 80\%, 100\%\}, and the number of classes remains unchanged for each data growth scenario. Here, a dataset with a small ratio is a subset of another dataset with a larger ratio.

\textbf{Comparison with state-of-the-art methods.} 
As shown in Table~\ref{tab:scenario_1}, our method achieves the best or comparable accuracy in all cases, suggesting its effectiveness. More critically, the computational cost (\ie MAdds) of our model is significantly lower than that of other methods, verifying that the proposed reward function enables the consideration of model efficiency. Specifically, our model achieves comparable accuracy with DEN (80.74 vs. 80.84) while the computational cost is much lower (252 M vs. 515 M). Similar phenomenons are widely observed in Table~\ref{tab:scenario_1}, indicating that our method achieves a better accuracy/computational efficiency trade-off than other state-of-the-art approaches.

\begin{table*}[t]
	\caption{\textbf{Scenario~II}: Comparison on ImageNet-1000 with different number of classes. We report Acc. (\%,$\uparrow$) and \#MAdds (Million, $\downarrow$).}
	\label{tab:scenario_2}
\newcommand{\tabincell}[2]{\begin{tabular}{@{}#1@{}}#2\end{tabular}}
 \begin{center}
 \begin{threeparttable}
    \resizebox{1.0\textwidth}{!}{
 	\begin{tabular}{lcc|cc|cc|cc|cc|cc|cc}\toprule
        \multirow{2}{*}{\tabincell{c}{Methods}}
        &\multicolumn{2}{c|}{\mata{10 classes}}&\multicolumn{2}{c|}{20 classes}&\multicolumn{2}{c|}{40 classes}&\multicolumn{2}{c|}{80 classes}&\multicolumn{2}{c|}{100 classes}&\multicolumn{2}{c|}{200 classes}&\multicolumn{2}{c}{1000 classes}\cr\cmidrule{2-3}\cmidrule{4-5}\cmidrule{6-7}\cmidrule{8-9}\cmidrule{10-11}\cmidrule{12-13}\cmidrule{14-15}
        & Acc. &MAdds& Acc. &MAdds&Acc.&MAdds&Acc&MAdds&Acc. &MAdds&Acc. &MAdds&Acc. & MAdds\cr
        \midrule
         MobileNetV2 &81.80	 	& 300 &85.10	 	& 300	& 81.10	& 300 &76.92	 	& 300	& 79.60	& 300 & 80.83 & 300 & 72.00 & \textbf{300}\\
         MobileNetV2 (1.4$\times$) &81.00 & 560 &85.70	 	& 560	& 81.30	&   560   &\textbf{77.85}	 	& 560 & 80.76	& 560  & 81.90 & 560  & 74.70 & 560  \\
         ResNet18 &\textbf{82.80} &1,814& 85.90 &	1,814 & 81.90	& 1,814  &75.60	& 1,814  & 79.30	& 1,814  & 79.89 & 1,814  & 72.12 & 1,814  \\	
         ResNet50 &69.60 &4,087 &81.80 & 4,087  & 78.85	& 4,087  &76.52	 	& 4,087 	& 80.30	& 4,087  & \textbf{82.89} & 4,087  & 77.15 & 4,087 \\
         \midrule
         MnasNet-A1 &80.60 & 323 &84.60 & 323  & 80.80	& 323  & 77.03 & 323  & 79.66 & 323  & 81.95 & 323  & 75.20 & 323 \\
         EfficientNet-B0 &81.40 &398 &86.00 & 398  & \textbf{82.10}	& 398  & 77.70 & 398  & 80.38 & 398  & 82.49 & 398  & 76.30 & 398 \\
         Meta-NAS &81.20 & 559 &85.50 & 559  & 80.75	& 559  & 75.03 & 559  & 77.48 & 559  & 80.53 & 559  & 74.30 & 559 \\
         \midrule
         \mata{D-EfficientNets} &81.00	 	& \textbf{145} &84.20 & 178 & 80.40	& 203 & 76.63 & 229 & 80.44 & 278 & 82.03 & 319 & 76.30 & 398 \\
         Progressive NN &81.20 & 149 &86.10 & 181  & 81.10	& 203  & 76.43 & 244  & 80.02 & 261  & 82.20 & 329  & 77.53 & 427  \\
         DEN &81.20 & 149 &86.10 & 258  & 81.35	& 315  & 77.60 & 439  & \textbf{80.84} & 515  & 82.09 & 549  & 72.99 & 672 \\
         \midrule
         \ourmethod~(ours) &81.20 & 149 & \textbf{86.40} & \textbf{176} & 81.90	& \textbf{195} & 77.68 & \textbf{242} & 80.52 & \textbf{257} & 82.12 & \textbf{293} & \textbf{78.13} & 395 \\
        \bottomrule
	\end{tabular}}
	 \end{threeparttable}
	 \end{center}
\end{table*}

\subsection{\blue{II: Growing Data with Increasing Label Space}}
In this section, we conduct experiments on a data growth scenario in which the label space is growing, \ie the new data have more classes than previous data. This scenario is more challenging since the distribution of new data may be remarkably different (\ie new classes) from that of previous data. The experimental datasets are constructed as follows.  

\textbf{Simulation of growing data:} 
In this experiment, we use the entire ImageNet-1000 to construct our growing datasets. Similar to Scenario~I, the dataset grows five times and each subset contains the first \{10, 20, 40, 80, 100, 200, 1000\} classes of the entire ImageNet, respectively. Similarly, the data of the latter case contain all the data of the former case.

\textbf{Comparison with state-of-the-art methods:} 
As shown in Table~\ref{tab:scenario_2}, our method is able to achieve comparable performance while requiring much lower computational cost.
Specifically, for ImageNet-1000, our method outperforms all the baseline methods in terms of accuracy. For ImageNet-100, our method outperforms manual-designed networks such as ResNet50, while requiring x15.7 fewer MAdds (257M vs. 4087M). Notably, when the number of classes is small, ResNet18 is significantly better than ResNet50. However, the situation turns around as the number of classes increases, which verifies our motivation that the optimal architecture may vary under different data distributions.

\subsection{Ablation Studies}

\begin{table}[t]
\newcommand{\tabincell}[2]{\begin{tabular}{@{}#1@{}}#2\end{tabular}}
\vspace{-0.1in}
\caption{\mata{Ablation studies on the adaptation condition. We report the accuracy of No-adjusted and Adjusted models on new current datasets $\mD_b\cup\mD_{s}$ and $\mD_b\cup\mD_{l}$, respectively.}
}
 \begin{center}
 \begin{threeparttable}
    \resizebox{\linewidth}{!}{
 	\begin{tabular}{c|c|cc}\toprule
 	Dataset & $H_{t}$ (Eqn.~\ref{eq:discrimination}) & No-adjusted (Acc. \%) & Adjusted (Acc. \%)\\
        \midrule
         $\mD_b\cup\mD_{s}$ & 0.65 & 64.90 & 65.04 (+0.14) \\
         \midrule
         $\mD_b\cup\mD_{l}$ & 7.76 & 72.54 & 73.58 (+1.04) \\
        \bottomrule 
	\end{tabular}
	}
	 \end{threeparttable}
	 \end{center}
	 \label{different_data}
	 \vspace{-0.2in}
\end{table}

\begin{table}[h]
\newcommand{\tabincell}[2]{\begin{tabular}{@{}#1@{}}#2\end{tabular}}
\caption{\mata{Comparison with nas-for-each (NFE) on ImageNet-100. NFE means ``search from scratch for each time of data growth''.}}
 \begin{center}
 \begin{threeparttable}
    \resizebox{\linewidth}{!}{
 	\begin{tabular}{c|c|cccc}\toprule
 	Metric & Method & 20\% data & 40\% data & 80\% data & 100\% data \\
 	\midrule
 	\multirow{2}{*}{\tabincell{c}{Acc. (\%)}}&
         NFE & 64.80 & \textbf{73.46} & 78.88 & 80.62  \\
         & AdaXpert (ours) & \textbf{64.90} & 73.28& \textbf{79.28}& \textbf{80.74}  \\
    \midrule
 	\multirow{2}{*}{\tabincell{c}{MAdds (M)}}&
          NFE & 294 &302 & 313 & 311   \\
         & AdaXpert (ours)& \textbf{171}& \textbf{199}& \textbf{232}& \textbf{252}  \\
    \midrule
     \multirow{2}{*}{\tabincell{c}{Search Cost\\ (GPU days)}} & NFE & 0.8 & 1.0 & 1.3 & 1.5  \\
 	& AdaXpert (ours) & \textbf{0.8}& \textbf{0.6}& \textbf{0.6}& \textbf{0.7}  \\
        \bottomrule
	\end{tabular}
	}
	 \end{threeparttable}
	 \end{center}
	 \label{tab:ablation_nfe_wd}
\end{table}

\textbf{Effectiveness of \rh{the adaptation condition in~Eqn.~(\ref{eq:discrimination})}}.
We conduct experiments to further demonstrate the effectiveness of our adaptation condition for architecture adjustment.
Specifically, we first prepare a base dataset $\mD_{b}$ (20\% training set of ImageNet-100 on Scenario~I) and a model trained on $\mD_{b}$. 
\rh{To demonstrate the necessity of architecture adaptation, we consider two different datasets,~\ie $\mD_{s}$ (with a small difference from $\mD_{b}$) and $\mD_{l}$ (with a large difference from $\mD_{b}$). To construct $\mD_{s}$,} we apply data augmetation techniques over the base dataset $\mD_b$. \rh{To construct $\mD_{l}$,} we use samples from another 20\% training set of ImageNet-100.

As shown in Table~\ref{different_data}, we report the accuracy differences (based on Eqn.~\ref{eq:discrimination}), adjusted/no-adjusted model accuracy on new current data $\mD_b\cup\mD_{s}$ and $\mD_b\cup\mD_{l}$, respectively. From the results, for similar new data, it is unnecessary to adjust the previous model architecture since the improvement is limited (\ie Acc: 64.90 vs 65.04). In contrast, adjusting the previous architecture for new different data is able to gain a larger performance improvement (\ie Acc: 72.54 vs 73.58). In this sense, it is important to use an adaptation condition to determine whether adjustments are needed for new data.

\begin{table*}[t]
    \centering
	\caption{Comparison of different architectures on ImageNet-1000. Our AdaXpert-\# architectures are searched on different subsets of ImageNet and then evaluated on the entire ImageNet dataset. "--" means unavailable results.}
	
	\begin{threeparttable}
	\resizebox{.895\textwidth}{!}{ 
    \begin{tabular}{lcccccc}
    \toprule
    \multicolumn{1}{c}{\multirow{2}[0]{*}{Architecture}} &
    \multicolumn{2}{c}{Test Accuracy (\%)} &
    \multicolumn{1}{c}{\multirow{2}[0]{*}{\#MAdds (M)}} &
    
    \multicolumn{1}{c}{Search Time} & 
    \multicolumn{1}{c}{\multirow{2}[0]{*}{Search Method}} &
    \multicolumn{1}{c}{\multirow{2}[0]{*}{Search Space}} \\
    \cline{2-3} & \multicolumn{1}{c}{Top-1} & \multicolumn{1}{c}{Top-5} & &  (GPU days) \\
    \midrule
    ResNet-18~\cite{resnet} & 69.8 & 89.1 & 1,814  & -- & \multirow{3}*{manual design} & \multirow{3}*{--} \\
     MobileNetV2 ($1.4\times$)~\cite{DBLP:conf/cvpr/SandlerHZZC18} & 74.7 & --  & 585  &  -- & \\
     ShuffleNetV2 ($2\times$)~\cite{ma2018shufflenet} & 73.7 & -- & 524 &  -- & \\
    \midrule
    NASNet-A~\cite{zoph2018learning} & 74.0 & 91.6 & 564  &  1,800 & RL-based & \multirow{2}*{NASNet} \\
     AmoebaNet-A~\cite{real2018regularized} & 74.5 & 92.0 & 555  &  3,150 & evolution &\\
    \midrule
    DARTS~\cite{liu2018darts} & 73.1 & 91.0 & 595   & 4 & gradient-based & \multirow{3}*{DARTS}\\
     P-DARTS~\cite{chen2019progressive} & 75.6 & 92.6 & 577   & 0.3 & gradient-based\\
     PC-DARTS~\cite{xu2020pcdarts} & 75.8 & 92.7 & 597 &  3.8 &gradient-based\\
    \midrule MobileNetV3-Large~\cite{howard2019searching} & 75.2 & -- & 219 & -- & RL-based & \multirow{10}*{Mobile Block} \\
     FBNet-C~\cite{DBLP:conf/cvpr/WuDZWSWTVJK19} & 74.9 & --  & 375   & 9 & gradient-based\\
     MnasNet-A3~\cite{tan2019mnasnet} & 76.7 & 93.3 & 403 & $\sim$3,791 & RL-based\\
     ProxylessNAS~\cite{cai2018proxylessnas} & 75.1 & 92.3  & 465   & 8.3 & gradient-based\\
     SPOS~\cite{oneshot_uniform} & 74.4 & 91.8 & 323  &  12 & evolution\\
     \mata{OFA-GPU}~\cite{Cai2020Once} & 76.4 & --  & 397 &  51.7 & evolution\\
     \mata{OFA-CPU}~\cite{Cai2020Once} & 78.7 & --  & 356 &  51.7 & evolution\\
     AtomNAS~\cite{Mei2020AtomNAS} & 75.9 & 92.0  & 367 & -- & gradient-based\\
     DNA-c~\cite{li2020block} & 77.8 & 93.7  & 466 & 25 & greedy search\\
     GreedyNAS-A~\cite{you2020greedynas} & 77.1 & 93.3  & 366 & 8 & greedy search \\
    \midrule
    \ourmethod-100 (ours) & 76.1 & 92.7  & 257 & 2.5  & \multirow{3}*{RL-based} & \multirow{3}*{Mobile Block}\\
    \ourmethod-200 (ours) & 77.1 & 93.3  & 293 & 3.5 & \\
    \ourmethod-1000 (ours) & 78.1 & 93.7  & 395 & 7 & \\
    \bottomrule
    \end{tabular}
    }
    \end{threeparttable}
	\label{tab:imagenet}
\end{table*}

\mata{
\textbf{Comparison with \emph{search from scratch}.} To further verify the superiority of \ourmethod, we also compare it with  ``search from scratch for each time data growth (namely NAS-for-Each, NFE)''. From Table~\ref{tab:ablation_nfe_wd}, our \ourmethod achieves better efficiency. At the 100\% data snapshot, the architecture performance obtained by \ourmethod is superior to NFE, this mainly benefits from the following two aspects: 1) The  AdaXpert exploits the previously learned knowledge to conduct the current learning. Similar ideas in GAN (e.g., Progressive GAN~\cite{karras2018progressive}) and NAS (e.g., PNAS~\cite{karras2018progressive} and CNAS~\cite{guo2020breaking}) have proven to be pretty effective. 2) The \ourmethod considers the different extent between current and previous data, and thus to adaptively control the computational cost of adjusted models. }

\subsection{Comparison on ImageNet-1000}

\begin{figure}[t]
    \centering
    \includegraphics[width=0.98\linewidth]{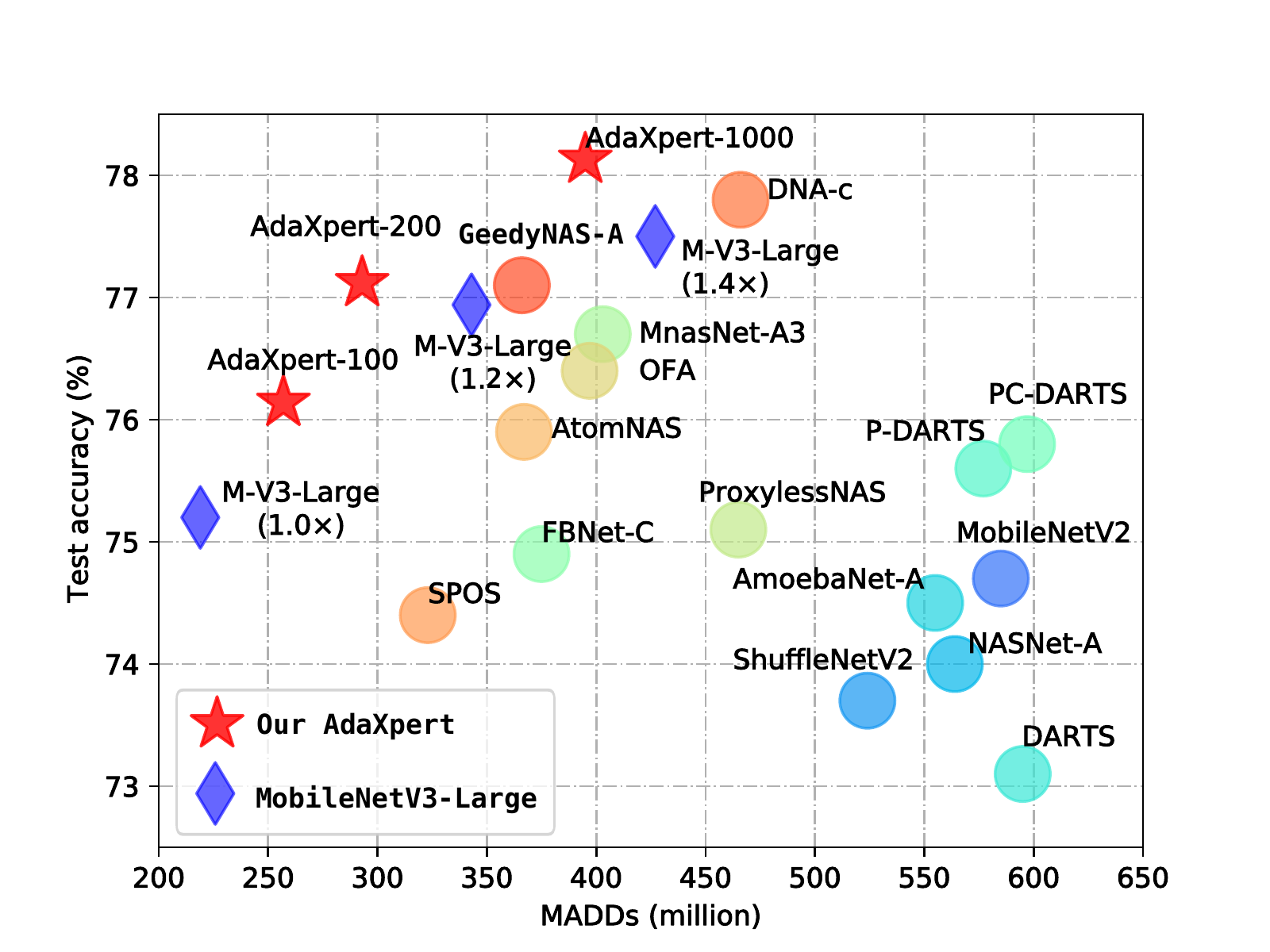}
    \vspace{-0.1in}
    \caption{Comparison between AdaXpert and state-of-the-art NAS methods on ImageNet. 
    `AdaXpert-\#' denotes our architecture searched on ImageNet-\#. 
    }
    \label{fig:imagenet}
    \vspace{-0.15in}
\end{figure}

\begin{figure}[t]
    \centering
    \includegraphics[width=0.48\textwidth]{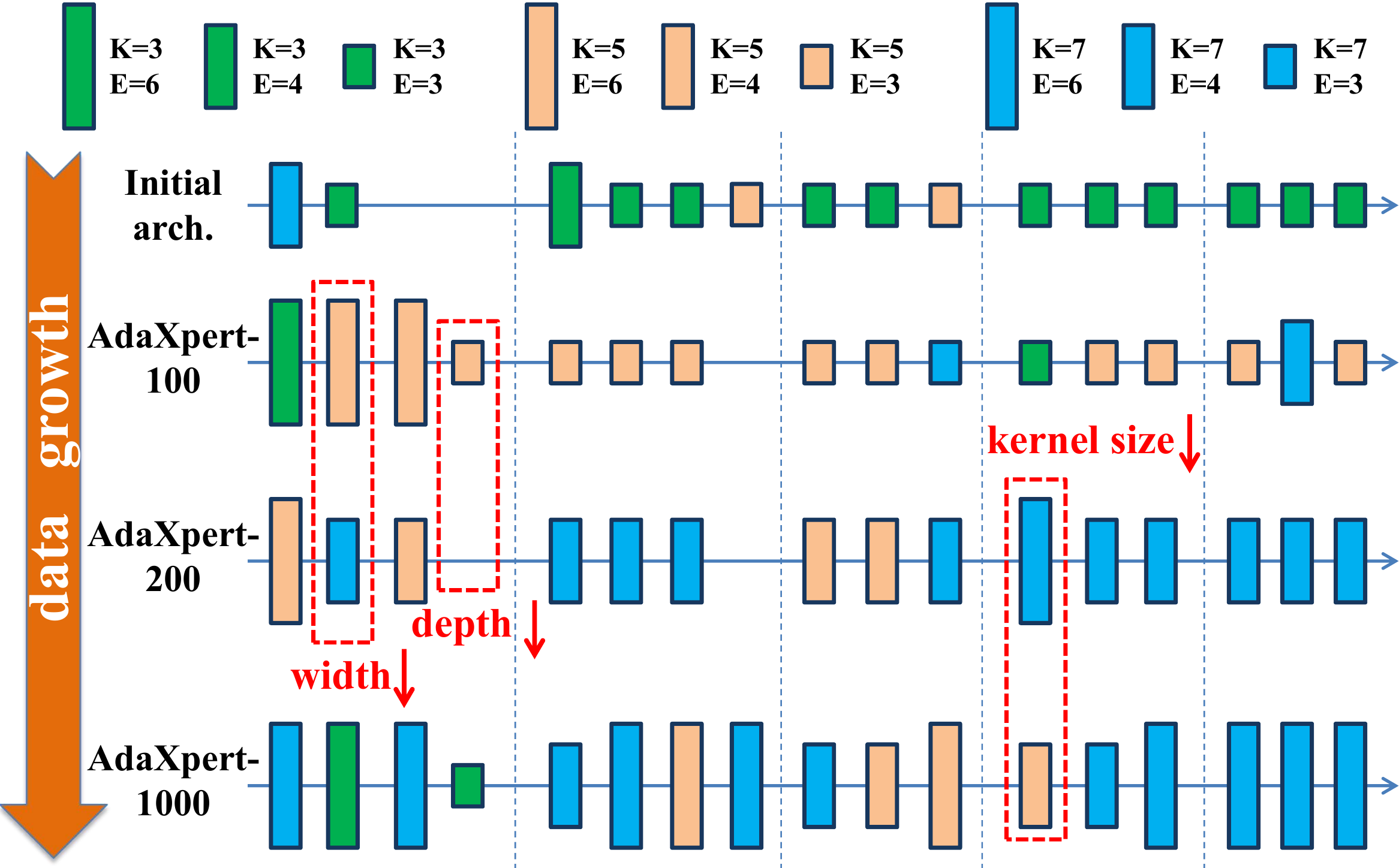}
    \caption{An illustration of the adjusted architectures of our \ourmethod. K and E denote kernel size and expansion ratio, respectively.}
    \label{fig:arch_diff}
\end{figure}

Our proposed method can also be regarded as a neural architecture search (NAS) method, which progressively searches for the optimal architecture on growing datasets. In this section, we compare our intermediate architectures, \ie \ourmethod-100, \ourmethod-200 and \ourmethod-1000, with existing NAS methods to further verify the effectiveness of our method. Here, \ourmethod-$\#$ is searched on ImageNet-$\#$. We re-train each \ourmethod-$\#$ model on the whole ImageNet-1000, as well as baseline methods.

As shown in Table~\ref{tab:imagenet} and Figure~\ref{fig:imagenet}, our AdaXpert-1000 achieves 78.1\% in terms of top-1 accuracy, 
\mata{which outperforms existing human-designed architectures and mostly considered state-of-the-art NAS models with different search spaces.} 
Surprisingly, our intermediate models, AdaXpert-100 and AdaXpert-200, also achieve comparable performance with most of \grammar{the} baseline methods in top-1 accuracy, but with fewer MAdds.
One of \grammar{the} possible reasons is that we exploit previous architectures and measure data distributions' difference of growing data, while general NAS methods search architectures from scratch. We also provide the visualization of our adjusted architectures in Figure~\ref{fig:arch_diff}. According to the data difference of each data growth, out \ourmethod adopts different architecture adjustment strategies. Although the model capacity generally increases as the data grow, it is observed that some redundant layers may be removed and some other layers' kernel size and expansion ratio may be reduced.

\section{Conclusion}
In this paper, we have proposed a new neural architecture adaptation method to efficiently adapt suitable neural architectures for growing data. Unlike existing methods \blue{that} neglect the knowledge from previous architectures, our method exploits the previous architecture and the data difference extent between current and previous data for achieving effective adaptation. Moreover, we have devised an adaptation condition to avoid unnecessary adjustments, thus further improving the network adaptation efficiency. Experimental results show that our method achieves state-of-the-art performance while enjoying less computational cost in two data growth scenarios (increasing data volume or number of classes). 
More critically, compared with the architectures searched on the entire ImageNet dataset by existing NAS methods, our architectures are able to achieve comparable accuracy/computational cost with fewer training data (\ie the subset of ImageNet).
In future work, it would be interesting to extend our method to adapt neural architectures for the growing data from diverse data domains. 

\noindent\textbf{Acknowledgements.}
\mata{This work was partially supported by the National Key R\&D Program of China (No. 2020AAA0106900), National Natural Science Foundation of China (NSFC) 62072190, Key-Area Research and Development Program of Guangdong Province (2018B010107001), Program for Guangdong Introducing Innovative and Enterpreneurial Teams 2017ZT07X183, Fundamental Research Funds for the Central Universities D2191240, Tencent AI Lab Rhino-Bird Focused Research Program (No. JR201902).}



\nocite{langley00}
\balance
\bibliography{example_paper}
\bibliographystyle{icml2021}
\end{document}


\onecolumn
\icmltitle{Supplementary Materials for \\ ``AdaXpert: Adapting Neural Architecture for Growing Data"}
\appendix



\icmlsetsymbol{equal}{*}

\begin{icmlauthorlist}
\icmlauthor{Shuaicheng Niu}{equal,scut,edu}
\icmlauthor{Jiaxiang Wu}{equal,tencent}
\icmlauthor{Guanghui Xu}{scut}
\icmlauthor{Yifan Zhang}{nus}\\ 
\icmlauthor{Yong Guo}{scut}
\icmlauthor{Peilin Zhao}{tencent}
\icmlauthor{Peng Wang}{npu}
\icmlauthor{Mingkui Tan}{scut,pz}
\end{icmlauthorlist}

\icmlaffiliation{scut}{School of Software Engineering, South China University of Technology, China}
\icmlaffiliation{pz}{Pazhou Laboratory, China}
\icmlaffiliation{tencent}{Tencent AI Lab, China}
\icmlaffiliation{nus}{National University of Singapore, Singapore}
\icmlaffiliation{npu}{Northwestern Polytechnical University, China}
\icmlaffiliation{edu}{Key Laboratory of Big Data and Intelligent Robot, Ministry of Education, China}

\icmlcorrespondingauthor{Mingkui Tan}{mingkuitan@scut.edu.cn}

\icmlkeywords{Machine Learning, ICML}
\vskip 0.3in 



\printAffiliationsAndNotice{\icmlEqualContribution} 

In the supplementary, we provide more implementation details and more experimental results of our proposed \ourmethod. We organize the supplementary material as follows.
\begin{itemize}[leftmargin=*]
    \item In Section~\ref{sec:adaxpert_details}, we provide more implementation details of our proposed \ourmethod.
    \item In Section~\ref{sec:eval_details}, we provide the evaluation details for all compared network architectures.
    \item In Section~\ref{sec:ada_condition}, we provide more experimental results for the threshold selection of our adaptation condition.
    \item In Section~\ref{sec:discussion_wd}, we demonstrate the effectiveness of Wasserstein distance on measuring the data difference.
    \item In Section~\ref{sec:lambda_sens}, we provide the sensitivity analysis \wrt the trade-off parameter $\lambda$ (in Eqn.~(5)).
\end{itemize}

\section{Implementation Details of \ourmethod}\label{sec:adaxpert_details}
During the whole data growth process, we maintain a supernet $\mN_t$ and a controller $\pi(\cdot;\theta_t)$. For each time data growth, we first fine-tune the previous supernet on current data $\mD_t$ to obtain $\mN_t$. After that, the controller $\pi(\cdot;\theta_t)$ is trained by using the evaluation signals provided by the supernet $\mN_t$. We introduce the training details of them as follows.   

\textbf{Supernet:} 
For growing Scenario I (in the main paper), we update the supernet 180 epochs for the 0.2 and 0.4 size of ImageNet-100, and we update the supernet 60 epochs for the 0.8 and 1.0 size of ImageNet-100. For growing Scenario II, we update the supernet 180 epochs for ImageNet-20 and ImageNet-40, and 60 epochs for others. Following~\cite{oneshot_uniform}, we train
the supernet by uniformly sampling sufficient architectures and train them sequentially. For each data growth, the supernet is fine-tuned with a learning rate of 0.045, a weight decay of $5\times 10^{-5}$ and a momentum of 0.9.

\textbf{Controller:}
The controller takes the previous architecture $\alpha_{t-1}$ and the difference extent $d_t$ (in Eqn.~(1)) between data as inputs, and then outputs an adjusted architecture $\alpha_t'$. In the following, we first introduce the architecture design of the controller and then its training details.  

For the controller design, we first use a two-layer fully
connected network (FCN) to extract features of the input architecture. Meanwhile, to represent the extent of data difference, following~\cite{pham2018efficient},
we build a learnable embedding vector for different $d_t$. We then concatenate the architecture embedding and data-difference embedding, and send them to the controller model. 
We adopt an LSTM to build the controller model. Since the architecture can be represented by a sequence of tokens~\cite{zoph2016neural,pham2018efficient}, the controller is able to adjust the network architecture by sequentially predicting the token sequences, including depth, width, and kernel size.
Here, we incorporate FCN parameters and the learnable embedding vectors into the parameters of the controller
and train them jointly. 

For each time the growth of data, we train the controller model for 6k iterations. 
We use Adam with a learning rate of $2\times 10^{-4}$ and a weight decay of $5\times 10^{-4}$ as the optimizer. We also add the controller’s sample entropy to the reward, which is weighted by $2\times 10^{-4}$.
The trade-off parameter $\lambda$ in Eqn.~(5) is set to $0.5\times 10^{-4}$ and $2.5\times 10^{-4}$ for Scenarios I and II, respectively. \mata{Here, the value of $\lambda$ is only tuned for the first adjustment (e.g., on ImageNet-20) and then fixed for all the subsequent adjustments. In this way, although the $\lambda$ for subsequent adjustments may not be optimal, the experimental results show that this has already achieved promising performance. We believe that a careful and efficient tuning of $\lambda$ may further improve the performance of AdaXpert, which we leave to our future work}. 

\section{Details of Architecture Evaluation}\label{sec:eval_details}
\textbf{Evaluation on ``subsets" of ImageNet-1000}. For fair comparisons, we train all architectures (including our \ourmethod) on the subset of ImageNet-1000 via the same setting and then test them on the corresponding test set. To be specific, we train each architecture for 180 epochs with a batch size of 256. We apply an SGD optimizer with a weight decay of $5\times 10^{-5}$ and a momentum of 0.9. Moreover, the learning rate starts with 0.1 and is divided by 10 at the 80 and 130 epoch. All architectures are evaluated using Tesla V100 GPUs.

\textbf{Evaluation on ``entire" ImageNet-1000}.
We report the performance of all compared methods on ImageNet-1000 according to their original papers. For \ourmethod models, we evaluate them using the evaluation methods provided by~\cite{lu2020neural}. Specifically, to accelerate the evaluation, we initialize our model weights with a pre-trained and publicly available once-for-all network~\cite{Cai2020Once} and then fine-tune it for 85 epochs. The fine-tune training adopts an RMSProp optimizer with a decay of 0.9 and momentum of 0.9. We set the batch normalization momentum to 0.99 and weight decay to 1e-5. We use a batch size of 512 and an initial learning rate of 0.012 that gradually reduces to zero via the cosine annealing schedule. The regularization settings are similar as in~\cite{tan2019efficientnet}: we use augmentation
policy~\cite{cubuk2020randaugment}, drop connect ratio 0.2, and dropout ratio 0.2.


\section{Threshold Selection for Adaptation Condition}
\label{sec:ada_condition}
In this section, we show more results on our adaptation condition (in Eqn.~(4)) to help algorithm engineers to choose a suitable threshold $\epsilon$ for determining the necessity of architecture adjustment. Same to the main paper, experiments are conducted on two considered scenarios, \ie increasing data volume (\textbf{Scenario I}) and the number of classes (\textbf{Scenario II}). 

For Scenario I, we report the value of $H_t$ (Eqn.~(4)) of ResNet18, MobileNetV2 and our \ourmethod that are well-trained on the 0.2 training set of ImageNet-100. The $H_t$ is then computed between the 0.2 and \{0.3, 0.4, 0.5, 0.6, 0.7, 0.8, 0.9, 1.0\} size of ImageNet-100. For Scenario II that label space is growing, we report the $H_t$ of the above models that are well-trained on ImageNet-20. Then, the $H_t$ is computed between the ImageNet-20 and ImageNet-\{30, 40, 50, 60, 70, 80, 90, 100\}. The \textbf{Left} and \textbf{Right} of Figure~\ref{fig:ht_illustration} show the results of Scenario I and II, respectively. With the growth of new data, the previous models suffer more severe accuracy difference. Based on these results, one can choose a suitable threshold to determine whether to adjust the model architecture, according to the task at hand. In this paper, we set the threshold $\epsilon$ to 0.02.

\mata{It worth mentioning that for the Scenario II of increasing label space, the model performance must degrade. However, in this case, the users still need this adaptation condition to measure whether the degradation greater than a given threshold $\epsilon$ (in Eqn. 4) and then determine whether to adjust.}

\begin{figure}[h]
\label{fig:ht_illustration}
\centering
\begin{minipage}[t]{0.3\textwidth}
\centering
    \includegraphics[height=0.75\textwidth]{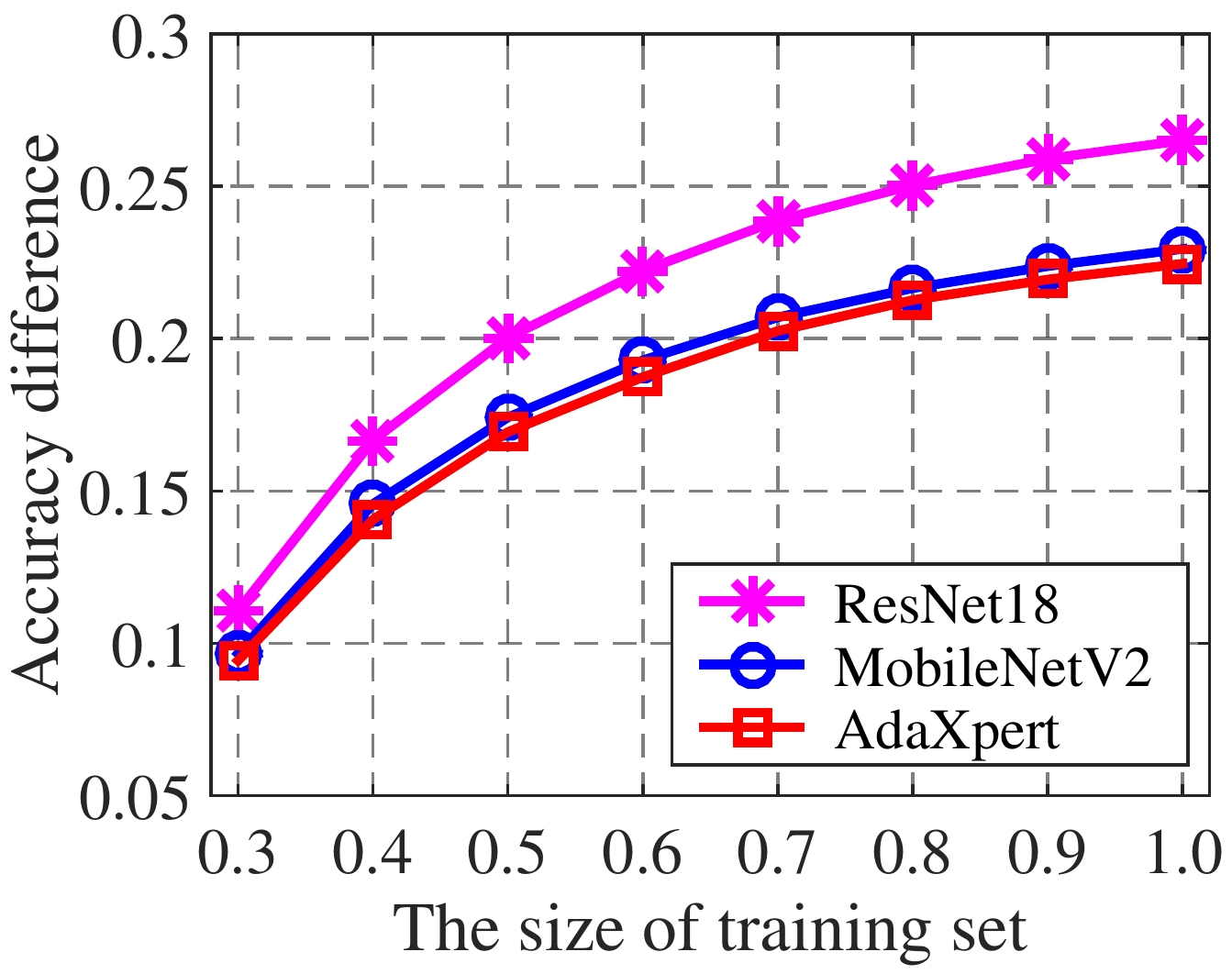}
\end{minipage}
\begin{minipage}[t]{0.3\textwidth}
\centering
    \includegraphics[height=0.75\textwidth]{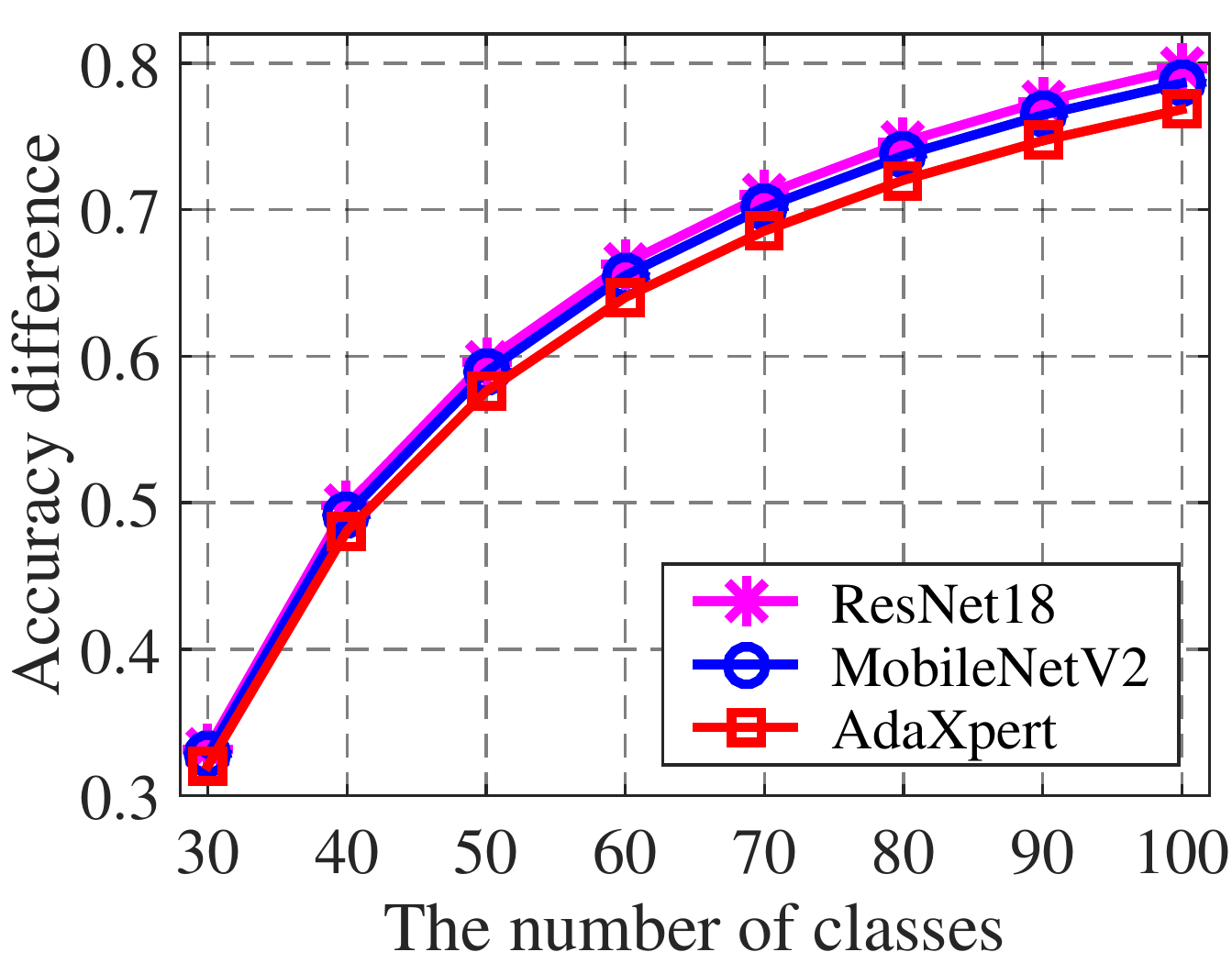}
\end{minipage}
\vspace{-0.1in}
\caption{An illustration of accuracy difference ($H_t$ in Eqn.~(4)) for different growing data. The \textbf{Left} and \textbf{Right} represent the increasing data volume and the number of classes, respectively.}
\label{fig:ht_illustration}
\end{figure}

\section{More Discussions on Wasserstein Distance}\label{sec:discussion_wd}

To compute distribution distance, one can use many metrics including Jensen-Shannon (JS) divergence~\cite{Fuglede2004Jensen},  Wasserstein distance (WD)~\cite{sriperumbudur2010non} and \etc~In this paper, \mata{we choose WD as our metric since 1) it is an effective metric to establish a geometric tool for comparing probability distributions, and it has been widely used in deep learning (e.g., GAN) and 2) it has a little stronger discrimination ability to recognize difference extents than JS divergence in our case.} In this section, we compare WD with JS on two considered scenarios, \ie increasing data volume (\textbf{Scenario I}) and the number of classes (\textbf{Scenario II}).
(1) For Scenario~I, based on our architecture searched on 0.2 training set of ImageNet-100, we compute the distance $d_t$ between 0.2 training set and \{0.4, 0.6, 0.8, 1.0\} training set. (2) For Scenario~II, based on our architecture obtained on ImageNet-20, we compute the distance between ImageNet-20 and ImageNet-\{40, 60, 80, 100, 200\}. 

\textbf{Computation details of WD and JS.}
As described in Sect.~3.2, we first feed current data $\mD_{t}$ and previous data $\mD_{t-1}$ into the previous model $\alpha_{t-1}$, and then obtain their corresponding feature matrices $\bM_{t}\in\mmR^{m\times q}$ and $\bM_{t-1}\in\mmR^{n\times q}$, respectively. Here, $m$ and $n$ denote the number of samples in $\mD_{t}$ and $\mD_{t-1}$ respectively, and $q$ denotes the feature dimension. We assume that $\bM_t$ and $\bM_{t-1}$ are from two multivariate Gaussian distributions $\mbP_t$ and $\mbP_{t-1}$, and use the Maximum Likelihood Estimation method to estimate its distribution parameters, \ie $\mbP_t\sim\mN(\mu_t,\Sigma_t)$ and $\mbP_{t-1}\sim\mN(\mu_{t-1},\Sigma_{t-1})$. Based on the above, the WD~\cite{takatsu2011wasserstein} between $\mD_t$ and $\mD_{t-1}$ is computed as follows:
\begin{align}
\label{eq:w_distance}
     \mW(\mD_t,\mD_{t-1}) = ||\mu_t-\mu_{t-1}||_2^2+   \text{tr}\Big(\Sigma_t\small{+}\Sigma_{t-1}\small{-}2(\Sigma_{t-1}^{1/2}\Sigma_t\Sigma_{t-1}^{1/2})^{1/2}\Big),
\end{align}
and the JS disvergence~\cite{Fuglede2004Jensen} is computed by:
\begin{align}
\label{eq:jsd}
     \text{JSD}(\mD_t,&\mD_{t-1}) = \frac{1}{2}\Big(\text{KL}(\text{P}_t || \frac{\text{P}_t\small{+}\text{P}_{t-1}}{2}) + \text{KL}(\text{P}_{t-1} || \frac{\text{P}_t\small{+}\text{P}_{t-1}}{2})\Big), 
      \\
      &\text{where}~~~ \text{KL}(\text{P}_t||\text{P}_{t-1})=\int_{-\infty}^{+\infty}\text{P}_t(x)\text{log}(\frac{\text{P}_t(x)}{\text{P}_{t-1}(x)})dx. \nonumber
\end{align}
Here, $\text{P}_t$ is the probability density function of $\mbP_t$. 

\begin{figure}[h]
\centering
\begin{minipage}[t]{0.32\textwidth}
\centering
    \includegraphics[height=0.75\textwidth]{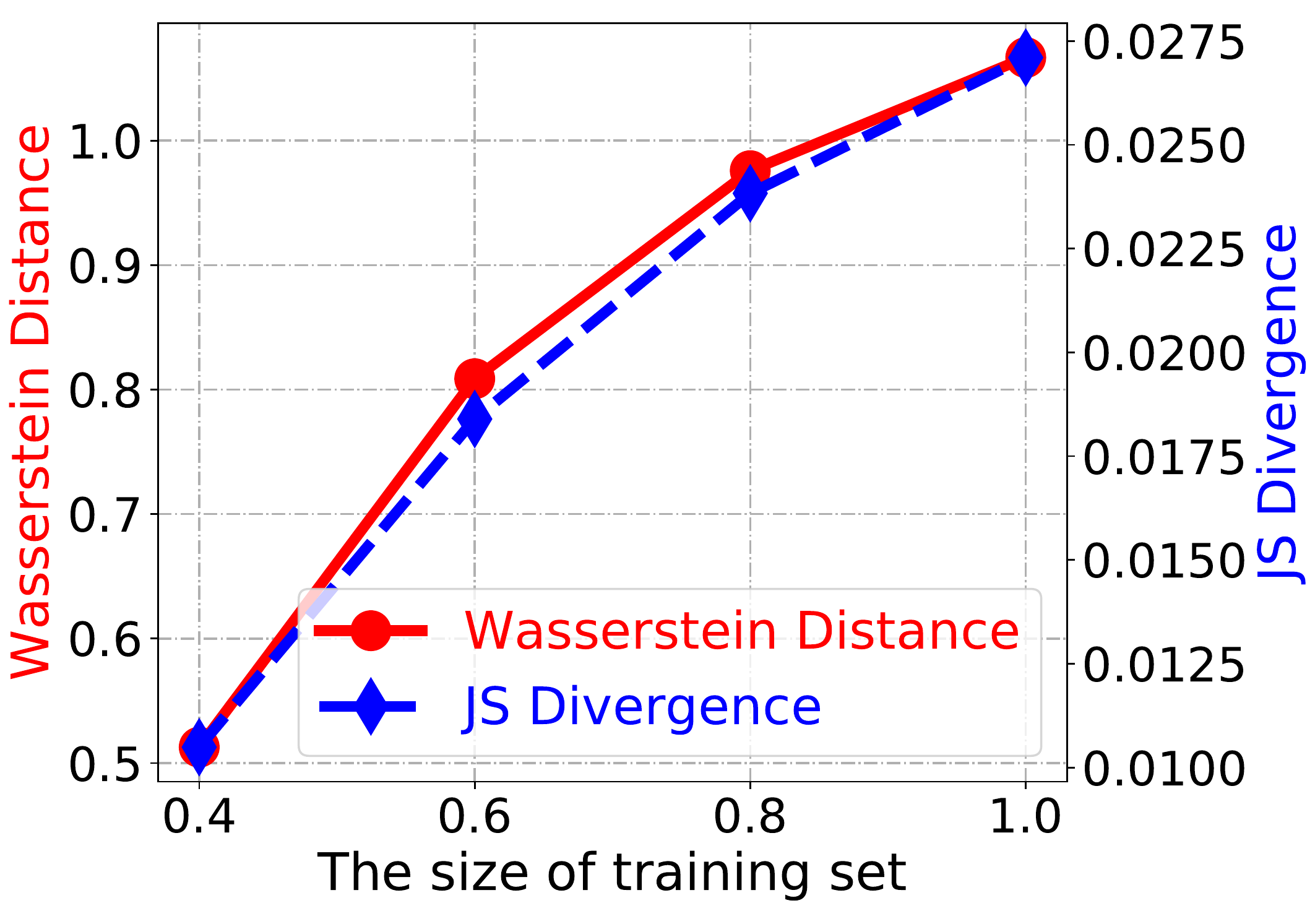}
    \label{fig:divergence_traningset}
\end{minipage}
\hspace{0.25in}
\begin{minipage}[t]{0.32\textwidth}
\centering
    \includegraphics[height=0.75\textwidth]{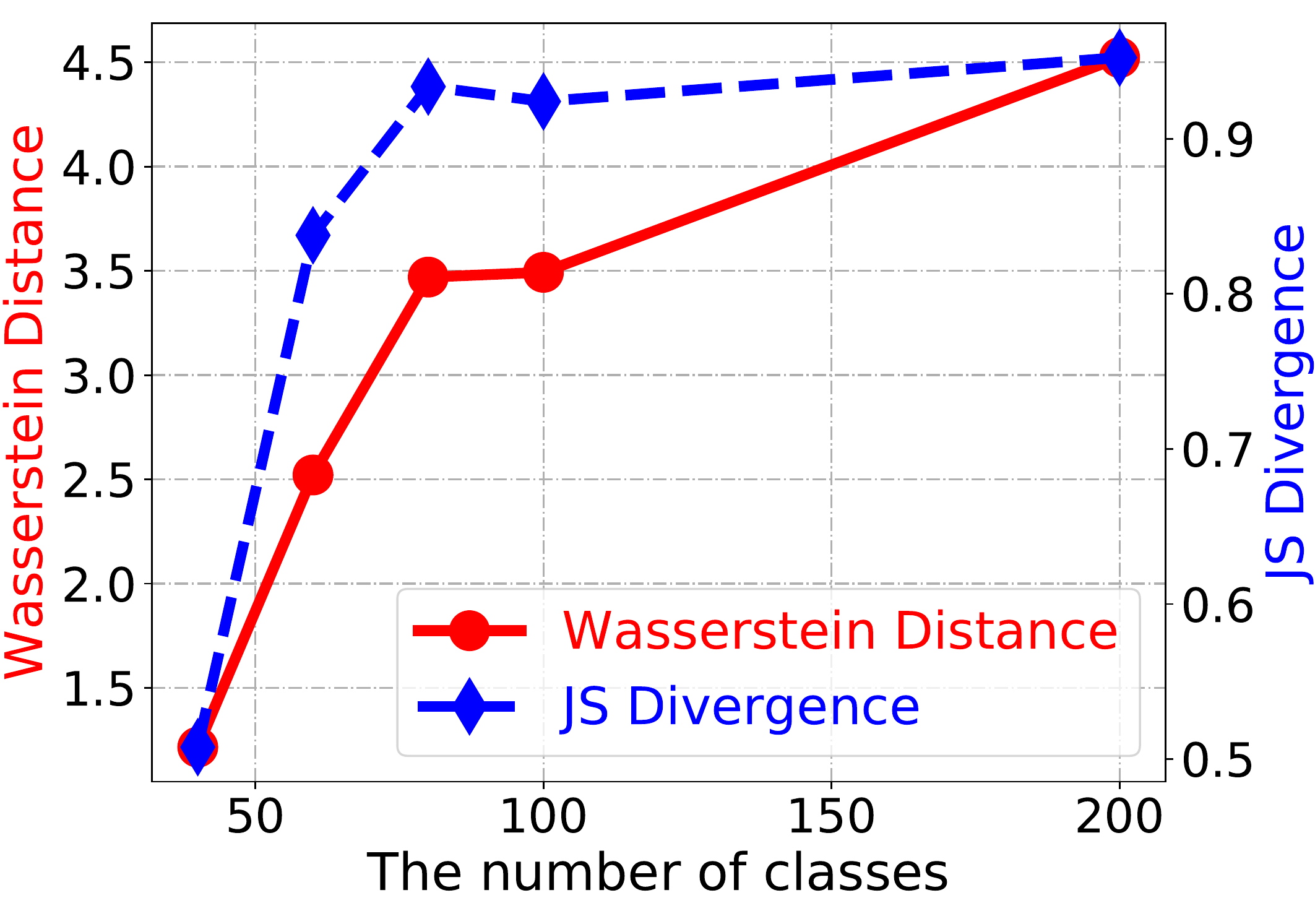}
    \label{fig:divergence_classes}
\end{minipage}
\caption{Comparison between WD and JS divergence.}
\label{fig:WD_JS}
\end{figure}

\textbf{Comparison between WD and JS.} 
As shown in Figure~\ref{fig:WD_JS}, both WD and JS are able to recognize the difference extent between current and previous data. In general, the more different the current data from previous data, the larger WD and JS are. For the Scenario II that label space is growing, WD show stronger discrimination ability than JS (see Figure~\ref{fig:WD_JS} (\textbf{right})). To be specific, the JS between ImageNet-20 and ImageNet-\{100, 200\} closes to 1 and thus fails to well recognize the difference extent between them. In contrast, the WD is still able to recognize the difference between ImageNet-100 and ImageNet-200.

\textbf{Effectiveness of Wasserstein Distance in \ourmethod.~}
In our method, we conduct different architecture adjustment strategies based on the Wasserstein distance (WD) between current and previous data. This enables our adjustment to achieve better model accuracy with moderate computational overhead. We compare our method with \ourmethod w/o WD, \ie the goal of reward function is only to maximize the model accuracy. Specifically, we conduct experiments on ImageNet-100 on Scenario~I. As shown in Table~\ref{tab:supp_ablation_wd}, with the data growth, `\ourmethod w/o WD' \blue{
is prone to} obtain a larger network for the current data. However, the network performance is comparable with our method in most cases. This further demonstrates the necessity of considering the data difference with WD to conduct the adjustment.

\begin{table}[h]
\newcommand{\tabincell}[2]{\begin{tabular}{@{}#1@{}}#2\end{tabular}}
\caption{Effectiveness of Wassertein distance in \ourmethod. `\ourmethod w/o WD' denotes \ourmethod without considering data difference (WD) in the reward function.}
 \begin{center}
 \begin{threeparttable}
    \resizebox{.55\linewidth}{!}{
 	\begin{tabular}{c|c|cccc}\toprule
 	Metric & Method & 20\% data & 40\% data & 80\% data & 100\% data \\
 	\midrule
 	\multirow{2}{*}{\tabincell{c}{Acc. (\%)}}
          &AdaXpert w/o WD & 65.12 & 73.76& 79.10& 80.96  \\
          &AdaXpert & 64.90 & 73.28& 79.28& 80.74  \\
    \midrule
 	\multirow{2}{*}{\tabincell{c}{MAdds (M)}}
 	&AdaXpert w/o WD & 279 &284 & 298 & 302   \\
 	&AdaXpert & 171 & 199 & 232 & 252   \\
        \bottomrule
	\end{tabular}
	}
	 \end{threeparttable}
	 \end{center}
	 \label{tab:supp_ablation_wd}
\end{table}

\begin{figure*}[h]
\centering
\begin{minipage}[t]{0.255\textwidth}
\centering
    \includegraphics[height=0.8\textwidth]{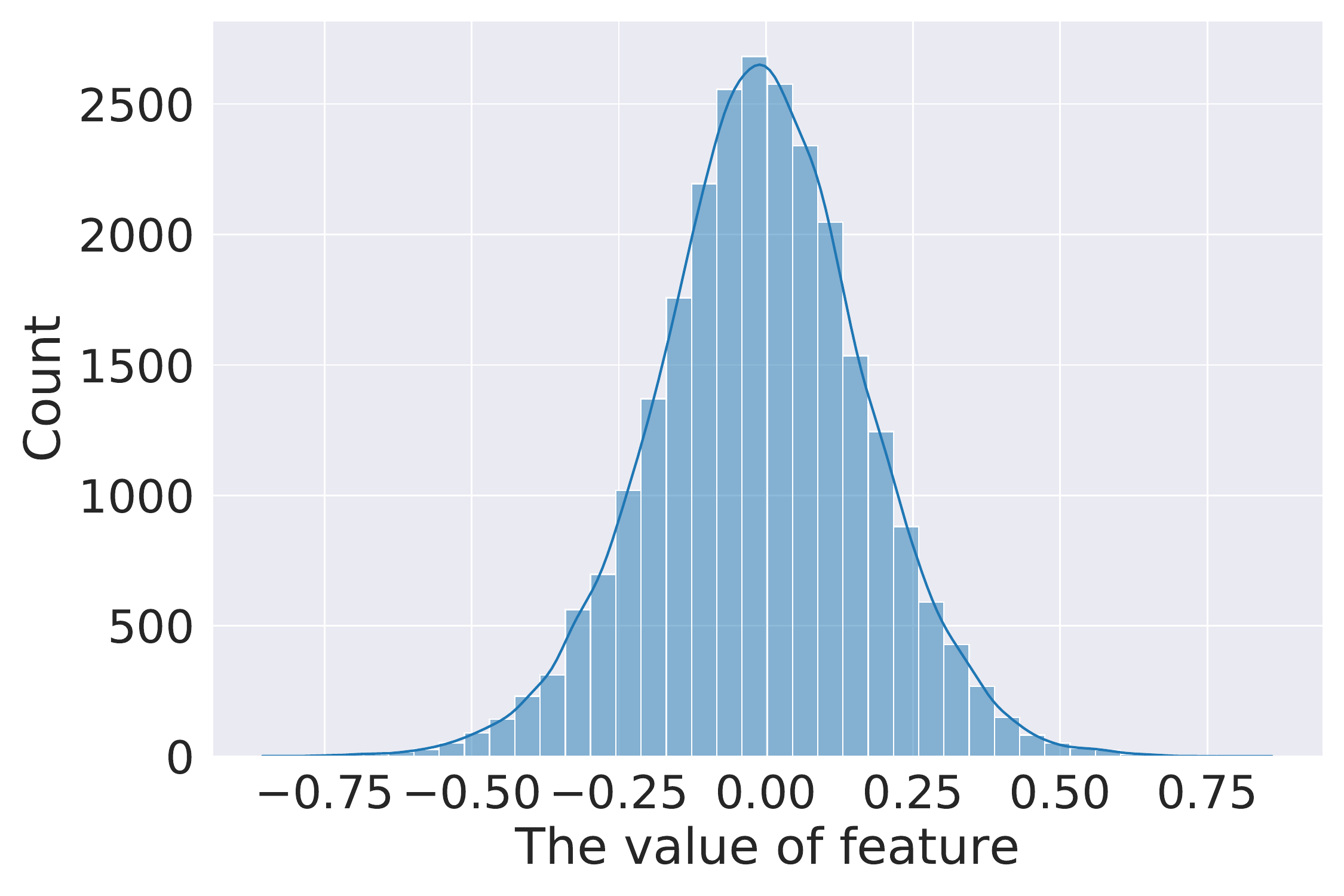}
\end{minipage}
\hspace{0.35in}
\begin{minipage}[t]{0.255\textwidth}
\centering
    \includegraphics[height=0.8\textwidth]{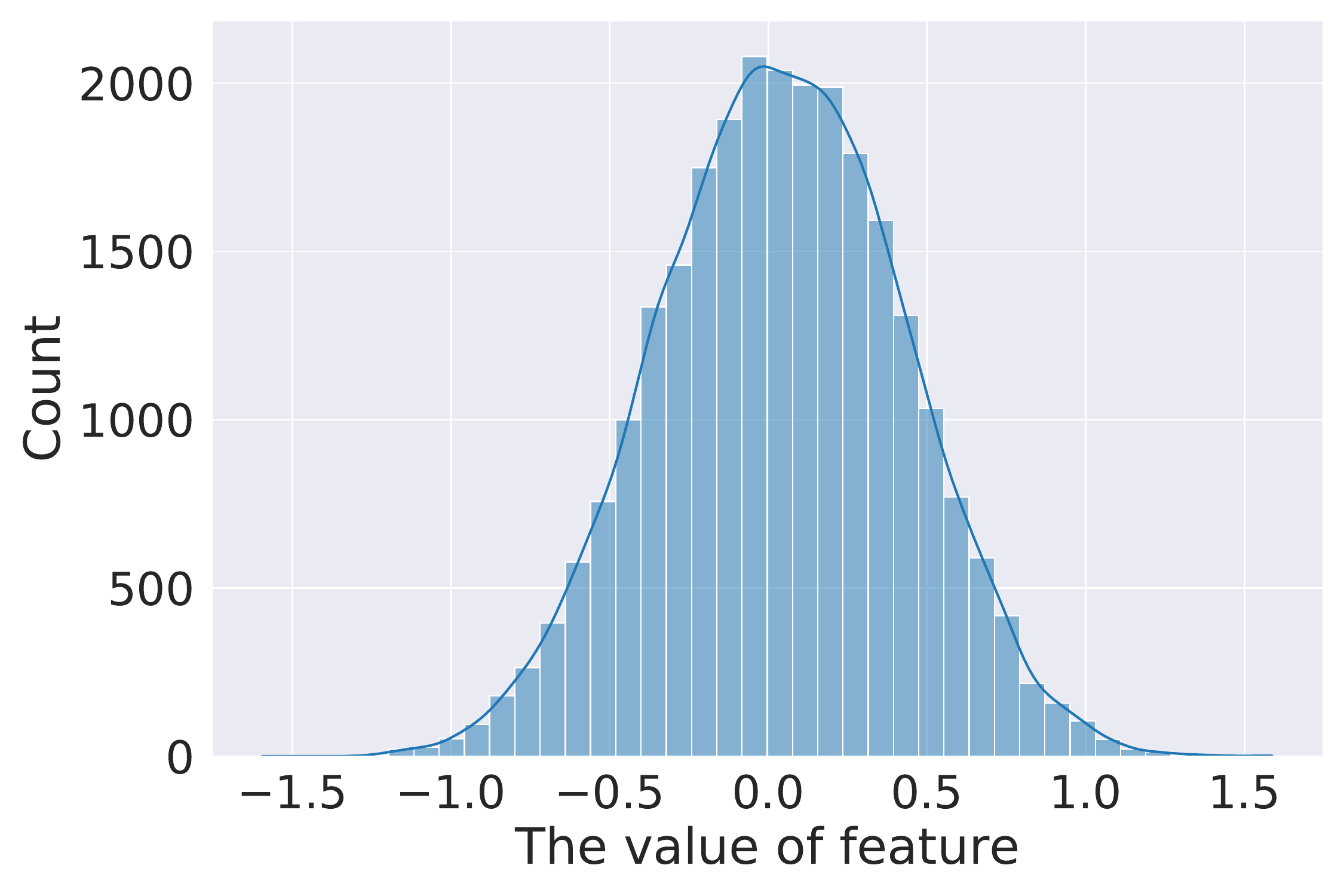}
\end{minipage}
\hspace{0.35in}
\begin{minipage}[t]{0.255\textwidth}
\centering
    \includegraphics[height=0.8\textwidth]{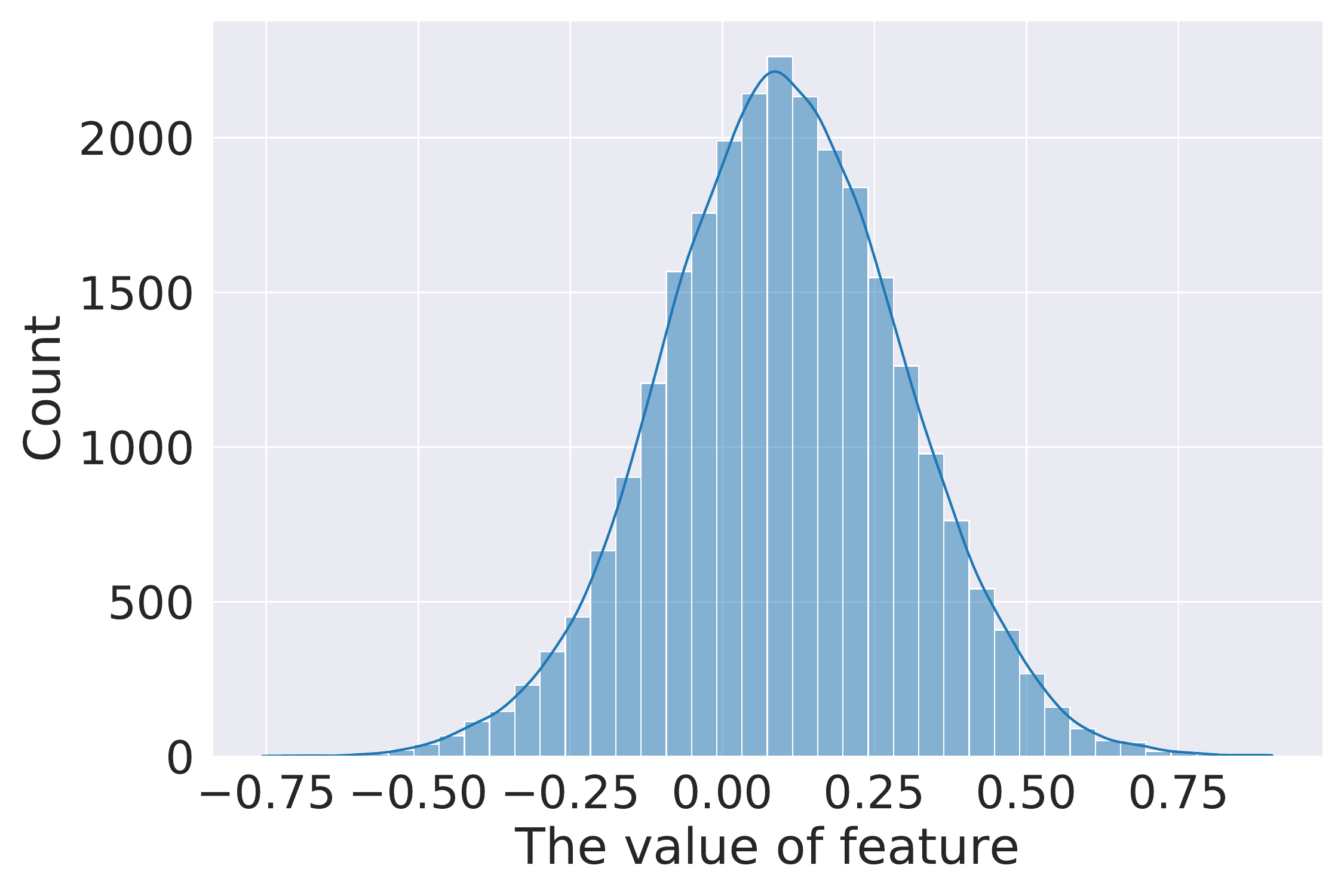}
\end{minipage}
\begin{minipage}[t]{0.255\textwidth}
\centering
    \includegraphics[height=0.8\textwidth]{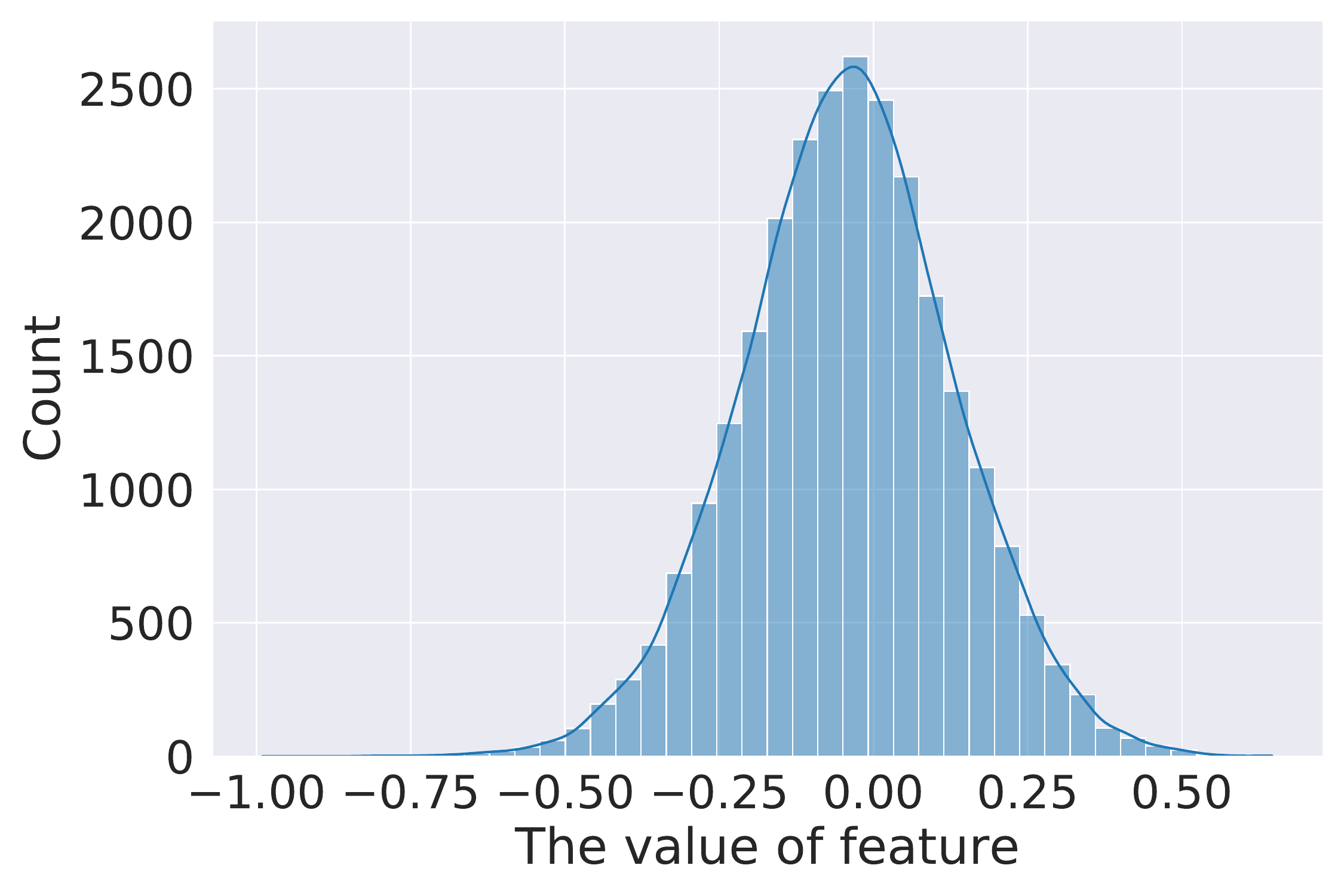}
\end{minipage}
\hspace{0.35in}
\begin{minipage}[t]{0.255\textwidth}
\centering
    \includegraphics[height=0.8\textwidth]{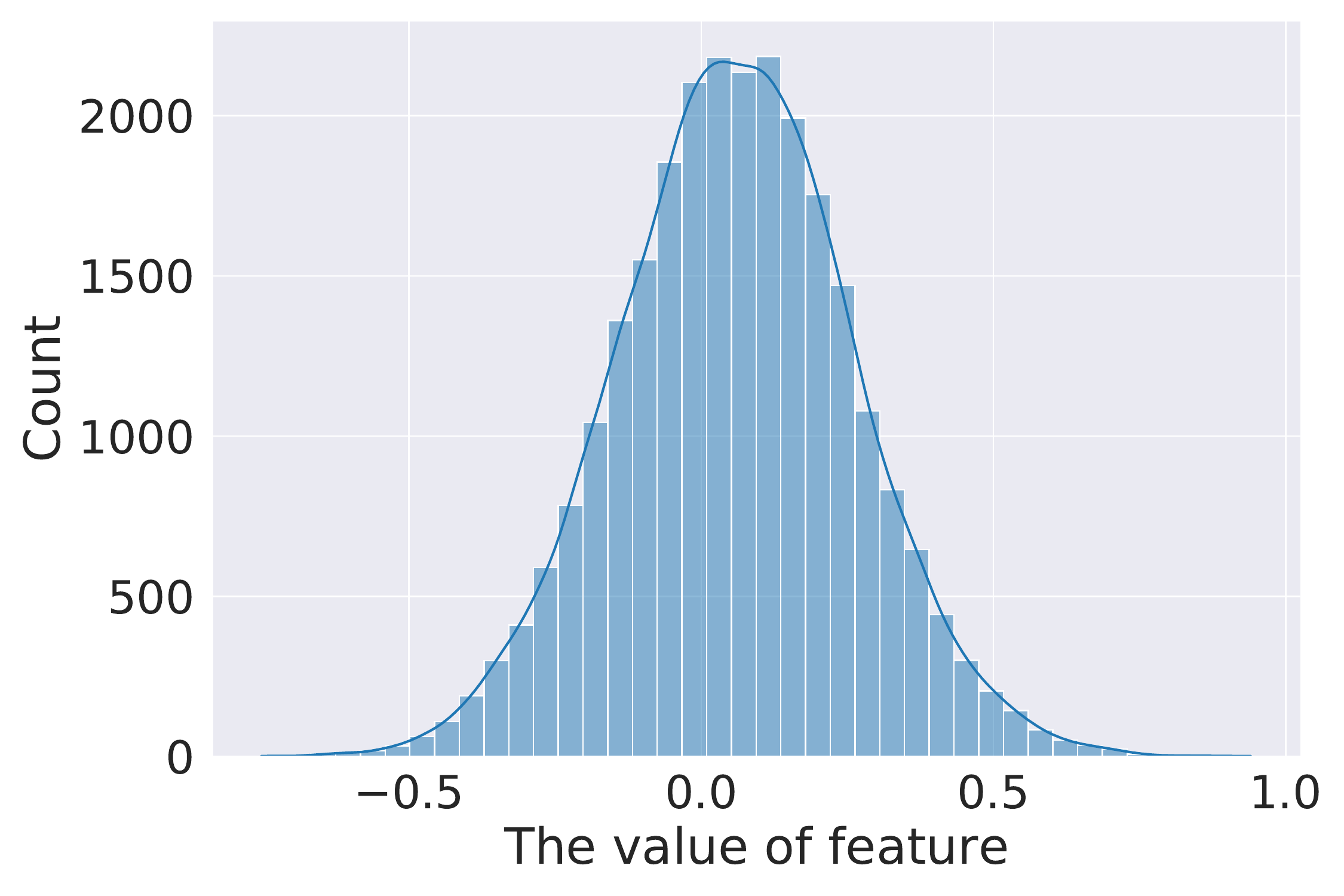}
\end{minipage}
\hspace{0.35in}
\begin{minipage}[t]{0.25\textwidth}
\centering
    \includegraphics[height=0.8\textwidth]{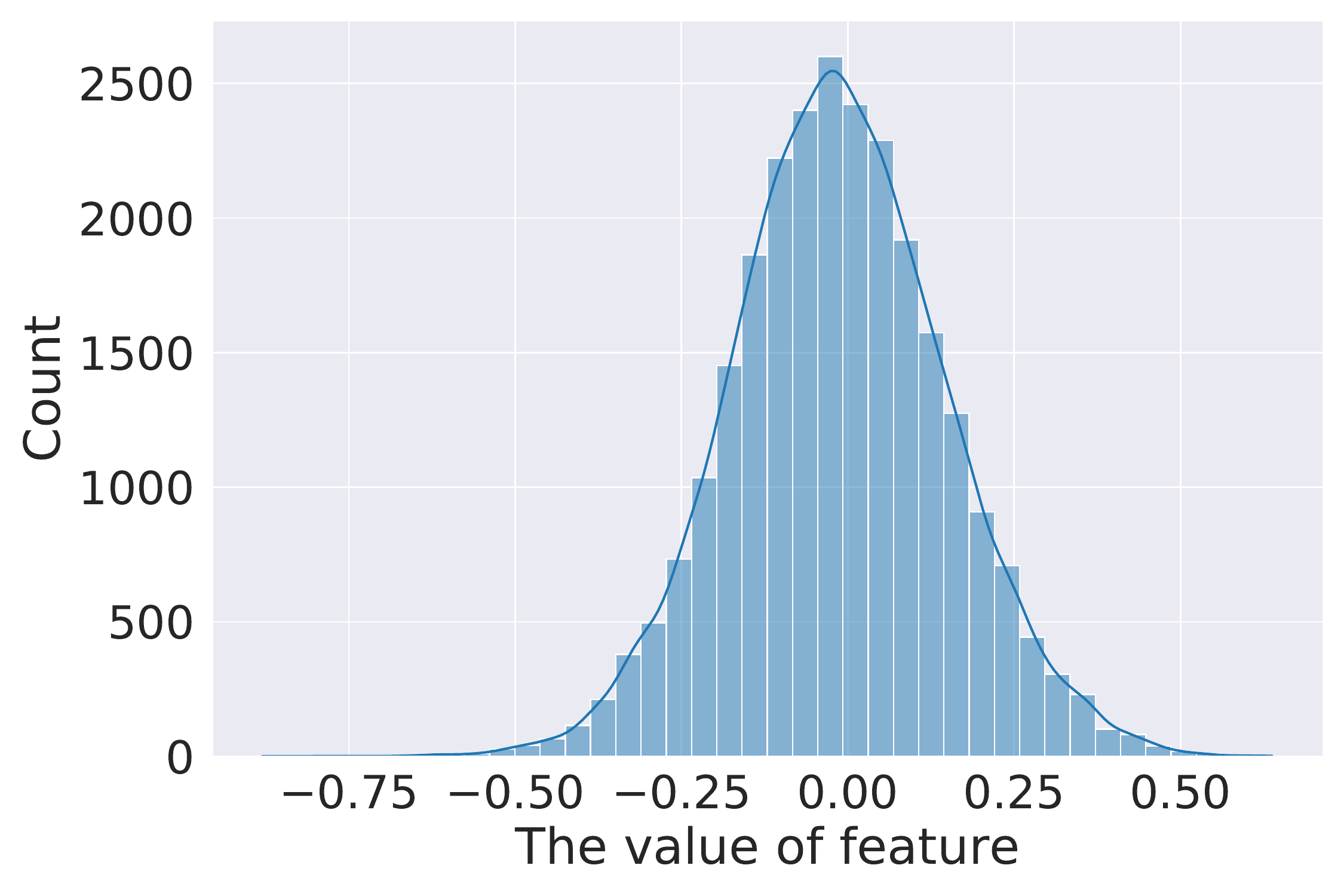}
\end{minipage}
\caption{Statistical histograms of data matrix on 6 randomly sampled dimensions.}
\label{fig:gaussian}
\end{figure*}

\textbf{Discussion on Gaussian Assumption.}
In this paper, we compute the WD between current and previous data by assuming that they are from two multivariate Gaussian distributions. Here, we empirically demonstrate the reasonability of this assumption. Based on our \ourmethod-20 that well-trained on ImageNet-20, we compute the sample matrix of ImageNet-40 (as described in Sect.~3.2) and randomly sample 6 dimensions to visualize its statistical histogram. As shown in Figure~\ref{fig:gaussian}, the sample features of each dimension approximately satisfy a Gaussian distribution. To achieve more accurate computation of WD, one can also use the non-parametric estimation methods~\cite{sriperumbudur2010non} as described in Sect.~3.2. 


\section{Parameter Sensitivity of $\lambda$ in Eqn.~(5)}\label{sec:lambda_sens}
\begin{figure}[h]
\centering
\begin{minipage}[t]{0.3\textwidth}
\centering
    \includegraphics[height=0.8\textwidth]{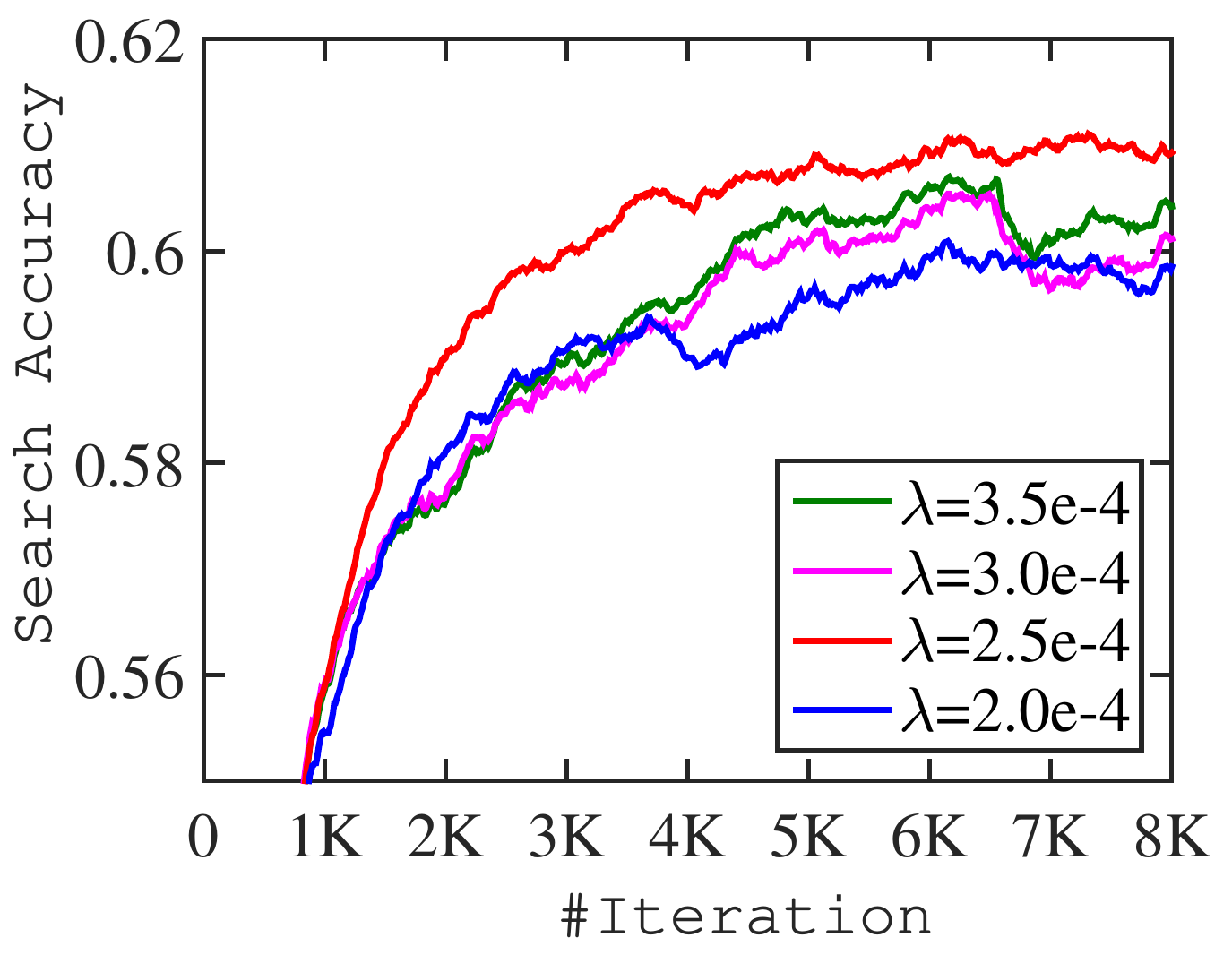}
    \label{fig:top1}
\end{minipage}
\hspace{0.1in}
\begin{minipage}[t]{0.3\textwidth}
\centering
    \includegraphics[height=0.8\textwidth]{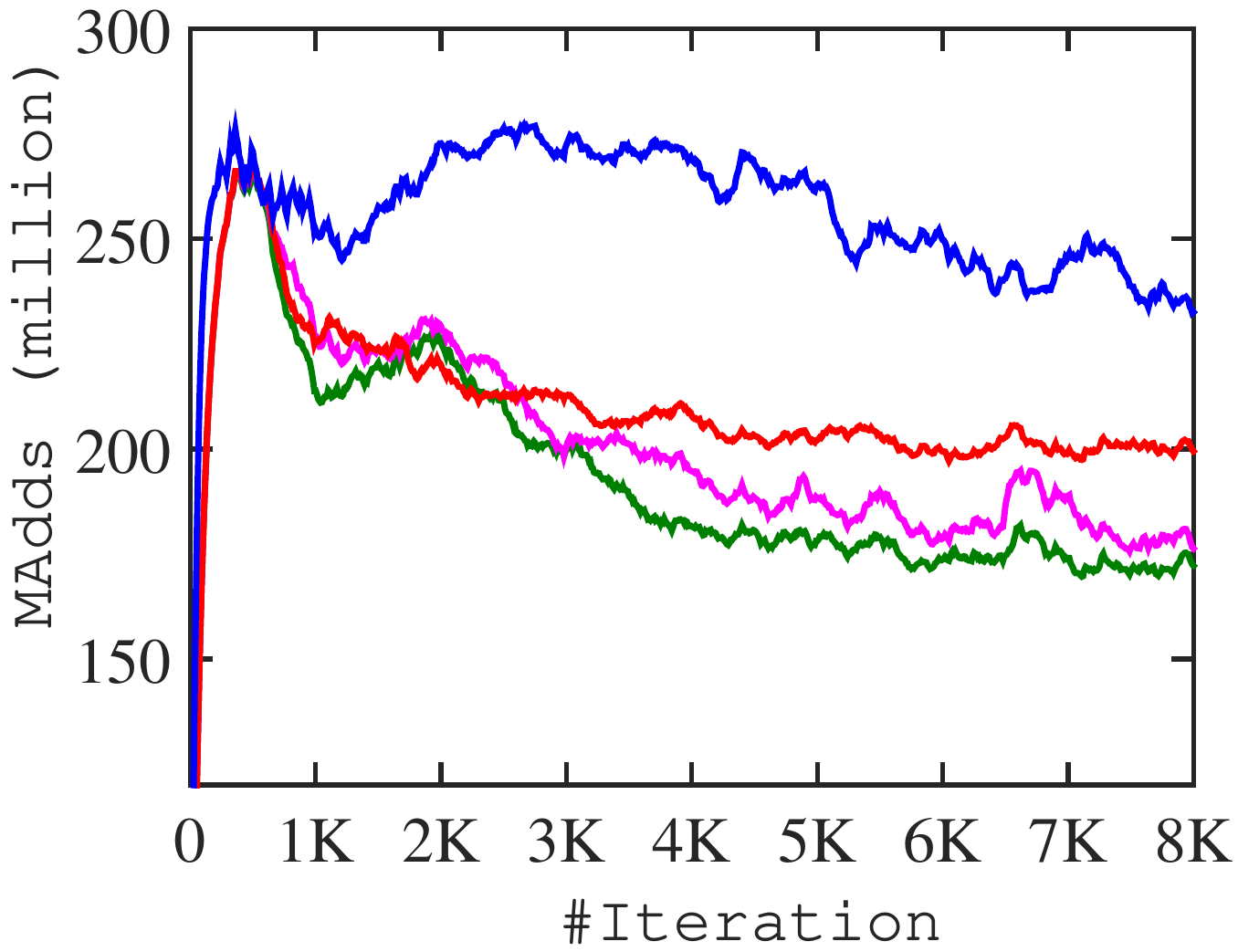}
    \label{fig:flops}
\end{minipage}
\caption{Training curves of our \ourmethod under different trade-off parameter $\lambda$.}
\label{fig:lambda_sens}
\end{figure}

In this section, we evaluate our method with different trade-off parameter $\lambda$ in Eqn.~(5) from $\{2.0, 2.5, 3.0, 3.5\}\text{e}^{-4}$. The experiments are conducted on adjusting \ourmethod-20 to \ourmethod-40. We report the validation accuracy and \#MAdds of the adjusted architectures in Figure~\ref{fig:lambda_sens} (\textbf{Left}) and (\textbf{Right}), respectively. 

From the results, with the increase of $\lambda$, our \ourmethod tends to find an architecture with fewer MAdds. However, the search accuracy (\ie validation accuracy) achieves the best when $\lambda=2.5\text{e}^{-4}$. Compared with $\lambda=2\text{e}^{-4}$, $\lambda=2.5\text{e}^{-4}$ achieves better search performance while with fewer \#MAdds. This result further demonstrates that a small model is able to achieve better accuracy than a large model in a certain dataset. In this sense, one can design well-performed architectures and meanwhile keep the model MAdds as few as possible.

\newpage
\bibliography{example_paper}
\bibliographystyle{icml2021}